\RequirePackage[hyphens]{url} 
\documentclass{article}

\usepackage[numbers, sort&compress]{natbib}
\usepackage{Definitions}

\usepackage{fullpage}
\usepackage{courier}
\usepackage{authblk}
\usepackage{calc}

\usepackage[utf8]{inputenc}
\usepackage[T1]{fontenc}
\usepackage[hidelinks]{hyperref}
\usepackage{xcolor}
\hypersetup{
    colorlinks,
    anchorcolor={black},
    linkcolor={black},
    citecolor={black},
    urlcolor={black}
}
\usepackage{booktabs}
\usepackage{amsfonts}
\usepackage{amssymb}
\usepackage{amsthm}
\usepackage{nicefrac}
\usepackage{microtype}
\usepackage{graphicx}
\usepackage[justification=centering]{subfig}
\usepackage{xcolor}
\usepackage{paralist}
\usepackage{mathtools}
\usepackage{enumitem}
\usepackage{xspace}
\usepackage[export]{adjustbox}
\usepackage{algorithm}
\usepackage[noend]{algpseudocode}
\usepackage{threeparttable}
\usepackage{multirow}
\usepackage{multicol}
\usepackage{floatpag}
\usepackage{color}
\usepackage{array}

\DeclareMathOperator{\dom}{\text{dom}}

\newcommand{\tnow}{t_{\text{now}}}

\newcommand{\tmax}{t_{\mathrm{max}}}

\title{Quantifying the Effects of Contact Tracing, Testing, and Containment Measures in the Presence of Infection Hotspots}

\author[1]{Lars Lorch}
\author[2]{Heiner Kremer}
\author[3]{William Trouleau}
\author[4]{Stratis Tsirtsis}
\author[5]{Aron Szanto}
\author[2,1]{\\Bernhard Sch{\"o}lkopf}
\author[4]{Manuel Gomez-Rodriguez}

\affil[1]{ETH Zürich\\ \href{mailto:lars.lorch@inf.ethz.ch}{lars.lorch@inf.ethz.ch}}
\affil[2]{Max Planck Institute for Intelligent Systems, \href{mailto:heiner.kremer@tuebingen.mpg.de,bs@tuebingen.mpg.de}{\{heiner.kremer,bs\}@tuebingen.mpg.de}}
\affil[3]{\'Ecole Polytechnique F\'ed\'erale de Lausanne, \href{mailto:william.trouleau@epfl.ch}{william.trouleau@epfl.ch}}
\affil[4]{Max Planck Institute for Software Systems,  
\href{mailto:stsirtsis@mpi-sws.org,manuelgr@mpi-sws.org}{\{stsirtsis,manuelgr\}@mpi-sws.org}}
\affil[5]{Zerobase Foundation, \href{mailto:aron@zerobase.io}{aron@zerobase.io}}

\newcommand\blfootnote[1]{%
  \begingroup
  \renewcommand\thefootnote{}\footnote{#1}%
  \addtocounter{footnote}{-1}%
  \endgroup
}

\begin{document}
\floatpagestyle{plain}

\maketitle


\begin{abstract}
Multiple lines of evidence strongly suggest that infection hotspots, where a single individual infects many others, play a key role in the transmission dynamics of COVID-19. 
However, most of the existing epidemiological models fail to capture this aspect by neither representing the sites visited by individuals explicitly nor characterizing disease transmission as a function of individual mobility patterns.
In this work, we introduce a temporal point process modeling framework that specifically represents visits to the sites where individuals get in contact and infect each other. Under our model, the number of infections caused by an infectious individual naturally emerges to be overdispersed. 
Using an efficient sampling algorithm, we demonstrate how to estimate the transmission rate of infectious individuals at the sites they visit and in their households using Bayesian optimization and longitudinal case data. 
Simulations using fine-grained and publicly available demographic data and site locations from Bern, Switzerland showcase the flexibility of our framework.
To facilitate research and analyses of other cities and regions, we release an open-source implementation of our framework.${}^*$
\end{abstract}

\section{Introduction}
\label{sec:introduction}
\hypersetup{urlcolor=blue}
\blfootnote{${}^*$Our code is publicly available at: \bfseries \url{https://github.com/covid19-model/}}
\hypersetup{urlcolor=black}
As countries around the world aim to counteract rising numbers of COVID-19 infections~\cite{covid-numbers-worldwide}, 
overwhelmingly growing evidence suggests that few infected people in infection 
hotspots, or \emph{superspreading events} (SSEs), may be responsible for both 
explosive early growth of cases and sustained transmission in later stages~\cite{adam2020clustering,carehomes,endo2020estimating,lau2020characterizing, frieden2020identifying,athreya2020effective}.
For example, in Hong Kong, the largest infection hotspots were traced back to four bars, which accounted for 32.5\% of all locally acquired infections from January 23 to April 28, 2020~\cite{adam2020clustering}.
In South Korea, an infection hotspot linked to a church was responsible for at 
least 60\% of all recorded cases by March 18, 2020, and over 1,000 
infections were traced back to a single individual~\cite{superspreader0}.
The first major outbreak in Germany occurred after an infected couple attended a carnival festivity in Heinsberg, with superspreading dynamics later verified by virus genome sequencing~\cite{walker2020genetic}.
These lines of evidence suggest that, for COVID-19, the number of infections caused by single infectious individuals is \emph{overdispersed}---most individuals infect few and a few infect many, exhibiting greater variance than expected under Poisson assumptions~\cite{cevik2020sars,hasan2020superspreading,zhang2020evaluating}.
Using carefully annotated tracing data, this has been identified as a root cause of SSEs~\cite{adam2020clustering, endo2020estimating,lau2020characterizing,  frieden2020identifying}.

Most of the existing epidemiological models for studying containment measures, including those developed and used in the context of the COVID-19 pandemic, neither explicitly represent sites of transmission, nor do they characterize exposures as a function of individual mobility patterns.
While this coarseness may be useful for fitting aggregate case trends, 
it makes conventional approaches unable to model the effects of granular interventions such as contact tracing or testing.
Moreover, existing models either assume or result in a Poisson distribution of infections caused by an infectious individual, also called \emph{secondary infections}, which fails to capture the high dispersion observed for COVID-19.\footnote{Overdispersion has also been observed in MERS and SARS~\cite{oh2015middle,lloyd2005superspreading,stein2011super}.}
As a result, these models have been of little use for identifying conditions under which hotspots emerge~\cite{ frieden2020identifying,cevik2020sars}, helping design and study control measures tailored to prevent SSEs~\cite{althouse2020stochasticity}, and predicting where infection hotspots are most likely to occur~\cite{zhang2020evaluating}.

In this work, we take a first step towards addressing the above limitations and present a data-driven framework for epidemiological modeling in the presence of overdispersed transmission dynamics and fine-grained containment measures.
Our main contributions are as follows:
\begin{itemize}
\item[\textit{(i)}] We introduce an event-based ``check-in'' mobility model that explicitly characterizes 
the frequency and duration of each individual'{}s visits to specific sites, which 
can be configured using a variety of publicly available data. 
\item[\textit{(ii)}] We develop a novel rate of transmission at sites that quantifies the influence of these individual mobility patterns as well as environmental drivers and containment measures on the risk of exposure that each infected individual poses to others at a site.
By using this transmission model and an explicit representation of the visited locations, our framework can directly characterize granular interventions that are targeted at particular sites and individuals (\eg, hygienic measures at work places, closures of schools, and contact tracing).
\item[\textit{(iii)}] We derive an efficient sampling algorithm for our model, which allows us to simulate the spread of COVID-19 under a variety of containment measures and counterfactual scenarios. 
Building on this procedure, we show how to estimate the disease transmission parameters using Bayesian optimization and longitudinal COVID-19 case data.
\end{itemize}
Our framework empirically scales to real-world cities and regions with hundreds of thousands of inhabitants and can be applied whenever simulated or real mobility traces as well as basic disease progression parameters (\eg, the incubation period and duration of infectiousness) are given.

We showcase our approach using fine-grained demographic data and site locations from Bern, Switzerland, and other regions in Germany and Switzerland.
Our results demonstrate that the number of individual disease transmissions---both overall and during a site visit---naturally 
emerges to be overdispersed, \ie, exhibiting higher variance than expected under the common Poisson assumption, and that our 
model is able to robustly characterize the observed COVID-19 case trends.
These findings hint at the potential of our framework as a complementary policy tool for studying the efficacy of containment measures, factors of disease transmission, and the nature of infection hotspots---hand in hand with existing societal and ethical considerations.
To facilitate research and analyses in this area, we release an open-source implementation of our framework~\cite{implementation-model}.

\section{Background} \label{sec:background}
Temporal point processes are random processes whose realizations $\Hcal = \{t_1, t_2, \dots, t_n\}$ consist of discrete events localized in time $t_i \in \RR^{+}$~\cite{aalen2008survival}. 
A temporal point process is commonly represented as a {\em counting process} $N(t)$, which counts the number of events that occurred before time $t$
\begin{align}\label{eq:background-counting-process-def}
    N(t) = \sum_{t_i \in \Hcal} u(t - t_i)
\end{align}
where  $u(x)$ is the unit step function and equals $1$ if $x \geq 0$ and $0$ otherwise.
Given a history of events $\Hcal(t) = \{t_i \in \Hcal \;|\; t_i \leq t \}$ until time $t$, we use a {\em conditional intensity function} $\lambda(t)$ to model the arrival probability of the next random event in the process.
More specifically, the conditional intensity function $\lambda(t)$ models the probability of an event occurring in an arbitrarily small time window after $t$.
We write
\begin{align}\label{eq:background-intensity-def}
    P(dN(t) = 1 \, | \, \Hcal(t)) = \lambda(t) \, dt
\end{align}
where the differential is defined as $dN(t) = N(t + dt) - N(t) \in \{0,1\}$. Here, $dt$ is an arbitrarily small time interval, and only one event can occur in $[t, t + dt)$.
The intensity function $\lambda(t)$ can be interpreted as an instantaneous rate of events per unit of time, for example, $\lambda(t)$~$=$~5 visits/week or $\lambda(t)$~$=$~1 infection caused/hour.
Note that $\lambda(t)$ may be time-varying and conditional on $\Hcal(t)$.

In stochastic differential equations (SDEs) with jumps,
the evolution of a set of state variables is characterized by the stochastic events of a set of counting processes.
Jump SDEs are commonly used for modeling dynamical systems with discrete stochastic events in continuous time, such as visits to sites or infections with a disease.
To illustrate, let $N(t)$ represent a counting process recording the number of emails sent to a person, and assume their inbox has a capacity of 1,000. Ignoring the deletion of emails, the change in the number of emails in the inbox $X(t)$ may be expressed by the SDE $dX(t) = u\big(1,000 - X(t)\big)dN(t)$, which increments $X(t)$ at every arrival of $N(t)$ until reaching the limit of 1,000.

\section{A Spatiotemporal Epidemic Model}
\label{sec:model}
In this section, our goal is to develop an agent-based, compartmental epidemiological model under which fine-grained spatiotemporal interventions can be expressed formally and the distribution of secondary infections 
induced by the model can exhibit overdispersion. 

To this end, our framework is composed of a collection of binary state variables that determine the mobility pattern, epidemiological 
condition, and testing status of each single individual $i \in \Vcal$.
We model the state transitions using stochastic differential equations (SDEs) with jumps, a model class that captures
{\em (i)} 
the stochastic nature of infection events and mobility patterns, 
{\em (ii) }
events in continuous time, i.e. \emph{not} in aggregate over a period, and 
{\em (iii) }
discrete state transitions---an individual either does \emph{or} does not get infected, visit a site, or get tested positively.

In the remainder of this section, we formally describe the dynamics of each state variable of the model. To ease the exposition, we distinguish between variables related to mobility, epidemiology, testing, and containment measures. 
Later, in Section~\ref{sec:sampling-and-estimation}, we then show how to generate random forward simulations of the entire model by devising an efficient sampling algorithm.
\begin{figure}[t]
    \centering
    \includegraphics[width=0.7\linewidth]{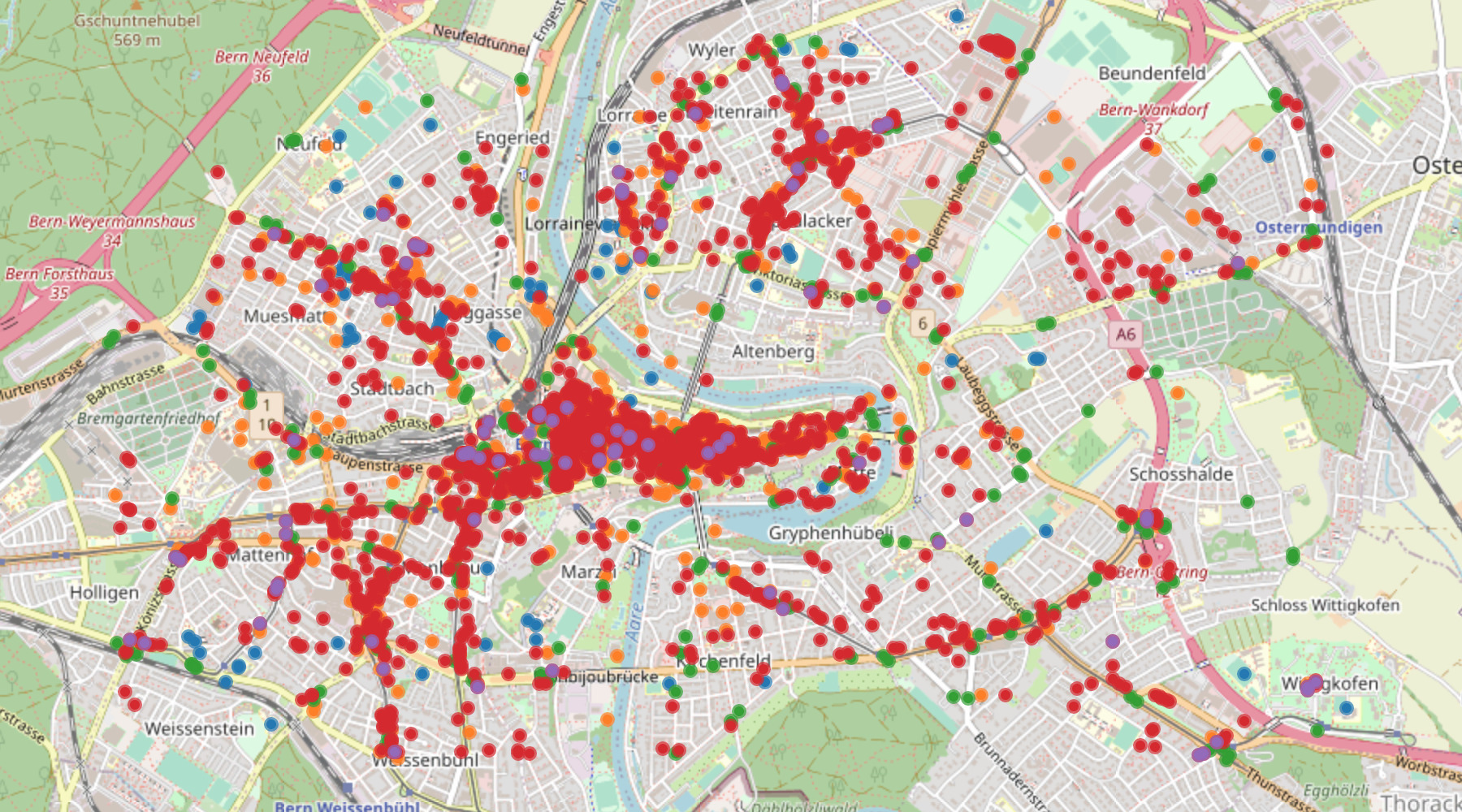}
    \caption{{\bf Site locations $\Scal$ by category in the mobility model of Bern, Switzerland.} The colored dots depict schools and 
research institutes (blue),  social places (orange), bus stops (green), workplaces (red), supermarkets (purple).
    }
    \label{fig:maps-main}
    \vspace{-3mm}
\end{figure}

\subsection{Mobility}\label{ssec:mobility}
For each individual $i \in \Vcal$ and a set of \emph{sites} $\Scal$ that individuals can visit, let $P_{i,k}(t) = 1$ if the individual is at site $k \in \Scal$ at time $t$ and 
$P_{i,k}(t) = 0$ otherwise. We characterize the value of the states $P_{i,k}(t)$ using the following SDE with jumps:
\begin{align} \label{eq:check-in}
dP_{i,k}(t) = dU_{i,k}(t) - dV_{i,k}(t)
\end{align}
$U_{i,k}(t)$ and $V_{i,k}(t)$ are counting processes that record the events of individual $i$ arriving at and leaving from site $k \in \Scal$, respectively. 
Thus, Equation (\ref{eq:check-in}) captures that $P_{i,k}(t)$ increments to $1$ after person $i$ arrives at site $k$ and decrements to $0$ after the person leaves.
We define the dynamics of the state transitions as 
\begin{equation} \label{eq:intensities-u-v}
\begin{split}
P\left( dU_{i,k}(t) = 1 \,|\, \mathcal{H}(t) \right) &= \eta_{i,k}(t) \,   \prod_{l \in \Scal} (1-P_{i,l}(t))  \, dt \\
P\left( dV_{i,k}(t) = 1 \,|\, \mathcal{H}(t) \right) &=  v_k\, U_{i,k}(t) \, dt 
\end{split}
\end{equation}
where $\eta_{i,k}(t)$ is the rate at which individual $i$ visits site $k$ and $1/v_k$ is the average duration of a visit to site $k$.
Equation (\ref{eq:intensities-u-v}) makes the state variable  $P_{i,k}(t) \in \{0,1\}$ well-defined by ensuring that $P_{i,k}(t) = 1$ for only one site $k \in \Scal$ at a time.

To configure the rates $\eta_{i,k}(t)$ and average duration $1/v_k$ for every individual and site, one can resort to publicly available data.
In our simulations, we configure the individual mobility statistics using the spatial distribution of real site locations, high-resolution population density data, country-specific information about household structure, and region-specific age demographics. 
We also assume that the probability that an individual $i$ visits a specific site $k$ decreases with the distance between their household and the site, similar to the gravity model~\cite{zipf1946p}.
As an example, Figure \ref{fig:maps-main} illustrates the sites $\Scal$ in an mobility model of Bern, Switzerland, which will be used for the case study in Section \ref{sec:results}.

\subsection{Epidemiology} \label{ssec:epidemiological-model}
To model the health status of each individual $i \in \Vcal$ while being in contact with others at sites $\Scal$ of the mobility model, we build on recent variations of the Susceptible-Exposed-Infected-Resistant (SEIR) compartment models that have been introduced in the context of COVID-19 modeling~\cite{li2020substantial,ferretti2020quantifying}.
More specifically, we define the epidemiological condition of each individual $i \in \Vcal$ using the indicator state variables 
$\mathbb{S}(t) = \{ S_i(t)$, $E_i(t)$, $I^a_i(t)$, $I^p_i(t)$, $I_i^s(t)$, $H_i(t)$, $R_i(t)$, $D_i(t) \}_{i \in \Vcal}$ with each $\in \{0, 1\}$,
whose meaning is specified in 
Table~\ref{tab:state-variables}.

\begin{table}[!t]
\caption{Epidemiological state variables $\mathbb{S}(t)$} \label{tab:state-variables}
\setlength{\tabcolsep}{4pt}
\centering
\begin{tabular}{ c l c c c}
  State & Description & Infected & Contagious & Symptoms \\\hline
 $S_i(t)$ &  is susceptible & \ - & - & - \\
 $E_i(t)$ &  is exposed & \ \checkmark & - & - \\
  $I^a_i(t)$ & is asymptomatic, &  &  &  \\
  &  ~~ mild course of disease  & \ \checkmark & \checkmark & - \\
  $I^p_i(t)$ & is presymptomatic,  &  &  & \\
  & ~~progresses to $I^s_i(t)$ later &\  \checkmark & \checkmark & - \\
  $I^s_i(t)$ & is symptomatic  & \ \checkmark & \checkmark & \checkmark \\
  $H_i(t) $ & is hospitalized & \ \checkmark & \checkmark & \checkmark \\
  $R_i(t) $ & is resistant and recovered & \ - & - & - \\
  $D_i(t) $ & has died & \ - & - & - \\
  \hline
\end{tabular}
\end{table}

\paragraph{Exposure}
First, we formally characterize the state transition of individual $i \in \Vcal$ from being susceptible ($S_i(t)$) to being exposed ($E_i(t)$) using the following jump SDEs: 
\begin{equation} \label{eq:seir-individual-expsoure}
\begin{split}
dS_i(t) &= - dN_i(t) \\
dE_i(t) &= dN_i(t) - dM_i(t) \\
\end{split}
\end{equation}
The counting process $N_i(t)$ models the exposure of individual $i \in \Vcal$ and thus forms the core component of our epidemiological model.
$M_i(t)$ models the subsequent transition to the asymptomatic or presymptomatic state.

To connect the epidemiological model to the mobility model, we assume that
{\em an individual's instantaneous rate of exposure increases by a constant site-specific transmission rate $\beta_k$ 
when in contact with another infectious individual at a site $k \in \Scal$.}
Consequently, the exposure rate of each individual $i$ only depends on the individual'{}s contacts at sites $k \in \Scal$ based on the mobility traces $P_{i,k}(t)$---and not the contacts of others.
We capture this model of exposure by the following conditional intensity function $\lambda_i(t)$ of the exposure counting process $N_i(t)$:
\begin{equation}
\hspace*{-5pt}
\lambda_i(t) = S_i(t)
\sum_{k \in \Scal} 
    P_{i,k}(t) 
\sum_{j \in \Vcal \backslash \{i\}} 
    \int_{t - \delta}^{t} K_{j,k}(\tau) ~ \gamma e^{-\gamma(t-\tau)} \, d\tau \label{eq:intensity-y}
\end{equation}
where 
\begin{equation*}
    K_{j,k}(\tau) = P_{j,k}(\tau) \, \beta_k \,  \big(
        I^{p}_j(\tau) + I^{s}_j(\tau) + \mu I^{a}_j(\tau)
    \big) 
\end{equation*}
and $P( dN_i(t) = 1 \,|\, \Hcal(t) ) = \lambda_i(t) \, dt$. In the above:
%
\begin{itemize}
    \item[\textit{(i)}] $\beta_k \geq 0$ is the base transmission rate of presymptomatic and symptomatic individuals at site $k$.
    Depending on the availability of labeled and unlabeled data, one may consider sharing the same parameter for all sites $k \in \Scal$ or for sites of the same category, \eg, distinguishing only indoors and outdoors.
    The scale $\mu \in [0, 1]$ denotes the relative transmission rate of asymptomatic compared to (pre-)symptomatic individuals.
    \item[\textit{(ii)}] $K_{j,k}(\tau) \in \{0, \mu \beta_k, \beta_k\}$ captures the effective contribution of individual $j$ to transmission at site $k$ and is non-zero if and only if $j$ is both infected (symptomatic, presymptomatic, or asymptomatic) and present at site $k$ at time $\tau$. 
    \item[\textit{(iii)}]
    $\int_{t - \delta}^{t} K_{j,k}(\tau) \, \gamma e^{-\gamma(t-\tau)} \, d\tau$
    models environmental transmission by accounting for the fact that the virus survives for some period of time $\delta$ on surfaces or in the air after an infected individual has left a site~\cite{vandoremalen2020aerosol}.
\end{itemize}
Thus, while traditional epidemiological models constrain the number of exposures to homogeneous Poisson distributions, our model in Equation (\ref{eq:intensity-y}) employs a stochastic and dynamically adjusting exposure rate for each individual $i \in \Vcal$ based on the mobility traces $P_{i,k}(t)$, under which this constraint is lifted. 
Infections within households can be characterized by adding an analogous term $\lambda_{\Hcal(i)}(t)$ with household transmission rate $\xi$ to $\lambda_i(t)$, which is outlined in Appendix \ref{app:household}.

\paragraph{Disease progression}
After an individual $i \in \Vcal$ got exposed, 
the subsequent state transitions are characterized by the counting processes
$W_i(t)$, $Y_i(t)$, $Z_i(t)$, $R^{a}_i(t)$ and $R^{s}_i(t)$. 
Let $a_i \sim \text{Bern}(\alpha_a)$ 
indicate whether an infected individual $i$ is asymptomatic or not.
Then, in the asymptomatic case where $a_i=1$, individual $i$ progresses from exposed to asymptomatic (via $M_i(t)$) and ultimately to recovered (via $R_i^a(t)$)
\begin{equation} \label{eq:seir-individual-asympt}
\begin{split}
dI^{a}_i(t) &= a_i  dM_i(t) - dR^{a}_i(t) 
\end{split}
\end{equation}
In the symptomatic case where $a_i=0$, the individual progresses from exposed to presymptomatic (via $M_i(t)$), from presymptomatic to symptomatic (via $W_i(t)$) and from symptomatic to resistant (via $R_i^s(t)$).
In addition, symptomatic individuals may be hospitalized or die from the disease. 
Let $h_i\sim \text{Bern}(\alpha_h)$ indicate whether the individual eventually requires hospitalization,
and $b_i \sim \text{Bern}(\alpha_b)$
whether they eventually die from the disease.
Then, a symptomatic individual may also transition from symptomatic infected to hospitalized (via $Y_i(t)$) and from symptomatic infected to dead (via $Z_i(t)$).
The presymptomatic ($I_i^p(t)$), symptomatic ($I_i^s(t)$) and resistant ($R_i(t)$) state variables are characterized by
\begin{equation} \label{eq:seir-individual-sympt}
\begin{split}
dI^{p}_i(t) &= (1-a_i)  dM_i(t) - dW_i(t) \\
dI^{s}_i(t) &= dW_i(t) - (1-b_i)  dR^{s}_i(t) - b_i  dZ_i(t) \\
dR_i(t) &= a_i dR^{a}_i(t) + (1-a_i) dR^{s}_i(t)
\end{split}
\end{equation}
Moreover, the hospitalized ($H_i(t)$) and dead ($D_i(t)$) states are given by
\begin{equation} \label{eq:seir-individual-recover}
\begin{split}
dH_i(t) &= h_i I^{s}_i(t) dY_i(t) - (1-b_i) H_i(t) dR^{s}_i(t) - b_i  H_i(t) dZ_i(t) \\
dD_i(t) &= b_i dZ_i(t)
\end{split}
\end{equation}
Since disease progression is disjoint from the mobility model, we follow the literature in modeling the above transition times using easy-to-sample log-normal distributions~\cite{lauer2020incubation,linton2020incubation}---starting at the time $E_i(t)$, $I^{p}_i(t)$, $I^{a}_i(t)$ or $I^{s}_i(t)$ become one, respectively, and terminating after their first event.
In practice, we fix their parameters based on estimates of the mean transition durations from the clinical COVID-19 literature.

\subsection{Testing} \label{ssec:testing-model}
Individuals are tested according to a testing policy $\pi_{\text{test}}(t)$, \eg, 
testing only symptomatic or vulnerable people, at a rate $\lambda_{\text{test}}(t)$, 
which can be chosen to match location-specific testing statistics. 
The test outcomes are only known after a reporting delay $\Delta_{\text{test}}$.
Formally, the counting process $T(t)$ records the number of known test outcomes by time $t$. Let $T^{+}_i(t)$ and $T^{-}_i(t)$ be the number of times an individual $i \in \Vcal$ has been tested positive and negative, respectively, 
by time $t$. Then, we characterize the state variables $T^{+}_i(t)$ and $T^{-}_i(t)$ 
using the following SDEs:
\begin{align}
dT^{+}_i(t) &= \big( E_i(t)+ I^{a}_i(t) + I^{p}_i(t) + I^{s}_i(t) \big) d_i(t) \, dT(t + \Delta_{\text{test}}) \nonumber  \\
dT^{-}_i(t) &= \big( S_i(t) + R_i(t) \big) d_i(t) \, dT(t + \Delta_{\text{test}})
\end{align}
where $d_i(t) \in \{0, 1\} \sim \pi_{\text{test}}(t)$ indicates whether $i$ is tested at time $t$ according to the policy.
In the above, a test result is positive if the individual is exposed ($E_i(t)$) or infected ($I_i^a(t)+I_i^p(t)+I_i^s(t)$), and negative if the individual is either susceptible ($S_i(t)$) or recovered ($R_i(t)$).
This can be relaxed to account for test specificity and sensitivity.

\subsection{Containment measures}\label{ssec:containment-measures}
In the above context, we can formally model a variety of containment measures that not only affect the broad population $\Vcal$ but also target specific sites or individuals, possibly in a time-variant fashion.
These may range from more granular (\eg, isolating individuals who have tested positive for 14 days or who had contact with a positively tested individual) to less granular (\eg, implementing a state of ``lockdown'' for the entire population).
The effect of mobility reduction and quarantine can be characterized by reducing the rates $\eta_{i,k}(t)$ at which individuals visit sites in the mobility model.
Hygienic measures (\eg, face masks) can be implemented by reducing the transmission rate $\beta_k$ at specific sites (\eg, work places).
In all cases, the measures reduce the conditional intensities $\lambda_i(t)$ of 
the exposure counting processes $N_i(t)$, possibly dynamically based on the values of other state variables at time $t$.

Moreover, if desired, we may assume that contacts between individuals at sites are registered by a peer-to-peer proximity-based tracing system, analogous to the smartphone-based Bluetooth systems that have been implemented in the context of the COVID-19 pandemic~\cite{covid-tracing-survey}.
A contact between individuals $i$ and $j$ will be registered if (i) their visit times at a specific site $k \in \Scal$ overlap, and (ii) both opt to use the proximity-based tracing system, \eg, by means of carrying a Bluetooth device.
Visit times are said to overlap when $P_{i,k}(t) = 1$ and $P_{j,k}(t) = 1$ for some site $k\in \Scal$ and time $t$.
When an individual $i$ is tested positive, their registered contacts may be advised to isolate or seek testing themselves as described in Section \ref{ssec:testing-model}.
For contact tracing, the type of intervention may depend on the risk of exposure caused by the positively tested individual, which can be estimated using our model. Appendix \ref{app:exposure-probability} provides further details.

\section{Model Simulation and Estimation}\label{sec:sampling-and-estimation}
\subsection{Epidemiological Sampling Algorithm}
\label{sec:sampling}
Having formally defined the model dynamics in Section \ref{sec:model}, we now introduce a procedure to sample trajectories of the individual epidemiological states $\mathbb{S}(t)$ over a time horizon $t \in [0, t_{\text{max}})$, which ultimately allows us to empirically study the spread of the disease under a variety of scenarios.
The initial conditions $\mathbb{S}(0)$, a testing policy $\pi_{\text{test}}(t)$, and the mobility traces $P_{i,k}(t)$ are assumed to be fixed a priori---from simulations of a synthetic mobility model as in Section \ref{ssec:mobility} or real-world data.

To sample a trajectory of the epidemiological state variables, we start by noticing that their values change at---and only change at---\emph{events} of the counting processes that model the transitions in the model SDEs.
Hence, all state variables $\mathbb{S}(t)$ are constant between two consecutive events when considering the event times of all counting processes in the model on \emph{one} timeline. This leads us to the backbone principle for generating random realizations of the model: 
we initialize the state variables $\mathbb{S}(0)$, sample the next time of state transition for each $i \in \Vcal$, and push these transition events onto \emph{one} temporally-sorted priority queue $Q$ that simultaneously tracks the next events for \emph{all} individuals in the model.
The algorithm then repeatedly loops through: 
{\em (i)} popping the next event $e$ from $Q$;
{\em (ii)} updating the state of individual $i$ associated with $e$;
{\em (iii)} sampling the next time $t$ of state transition $e'$ for $i$;
and {\em (iv)} pushing $e'$ to $Q$ with priority $t$.
As explained in Section \ref{sec:model}, we fix the time-to-event distributions of all processes {\em not} concerning exposure, \ie, excluding $ \{ N_i(t) \}_{i \in \Vcal}$, to independent, easy-to-sample distributions as estimated by clinical COVID-19 literature.
This means that once an individual is exposed, sampling the following times of state transition, \eg, to symptomatic and recovered states, is trivial.
However, sampling the time of exposure of $i$, \ie, the first event time of $N_i(t)$, is hard because the rate $\lambda_i(t)$ dynamically interacts with all other stochastic state variables $\mathbb{S}(t)$ via the mobility model $P_{i,k}(t)$.
To be able to sample from $N_i(t)$, we decompose the intensity $\lambda_i(t)$ into a sum of contributions $\lambda_{j \rightarrow i}(t)$ caused by other individuals~$j$:
\begin{equation}
\lambda_i(t) = S_i(t) \sum_{j \in \Vcal \backslash \{i\}} \sum_{k \in \Scal} P_{i,k}(t) \, \int_{t - \delta}^{t} K_{j,k}(\tau) \gamma e^{-\gamma(t-\tau)} \, d\tau \quad =: \quad S_i(t) \sum_{j \in \Vcal \backslash \{i\}} \lambda_{j \rightarrow i}(t) \label{eq:app-individual-exposure}
\end{equation}
where the last summation over $j \in \Vcal \backslash \{i\}$ is sparse as it only indexes over contacts of individual $i$ after time $t$.
Note that $\lambda_{j \rightarrow i}(t) = 0$ when $i$ and $j$ are not in contact directly or when $j$ left site $k \in \Scal$ more than $\delta$-time before $i$ arrived.
By Equation (\ref{eq:app-individual-exposure}), if individual $i$ is susceptible, the counting process $N_i(t)$ can be seen as a superposition of several processes $N_{j \rightarrow i}(t)$ with intensities $\lambda_{j \rightarrow i}(t)$.
This implies that the time-to-event distribution of $N_i(t)$ is equivalent to the distribution of the time to the \emph{first} arrival of all processes $N_{j \rightarrow i}(t)$ \cite{aalen2008survival,de2019temporal}.
Using the temporal ordering invariant of $Q$, we can thus process valid exposure events on the fly. 
Whenever an individual $j$ becomes infectious, \ie, 
$I^a_j = 1$ or $I^p_j = 1$, we sample the next exposure event that $j$ \emph{causes} for every individual $i$ in contact with $j$ in the future at rate $\lambda_{j \rightarrow i}(t)$, and push these events onto $Q$.
Later, when an exposure event $e$ for individual $i$ is popped from $Q$ in step {\em (i)}, we check whether $e$ is the \textit{first} exposure of $i$ by verifying $S_i(t) = 1$, and discard subsequent exposure events for $i$.

To sample the next event time of the subprocess $N_{j \rightarrow i}(t)$ after time $t'$, we use the principle of {\em thinning}~\cite{de2019temporal}. We can generate a valid sample from $N_{j \rightarrow i}(t)$ by repeatedly adding $\tau \sim \text{Expo} (\lambda^{\text{max}}_{j \rightarrow i} ) $ to $t'$ and stopping with probability
$\lambda_{j \rightarrow i} (t') /\lambda^{\text{max}}_{j \rightarrow i} $ at a given iteration, 
where $\lambda^{\text{max}}_{j \rightarrow i}$ is an upper bound on $\lambda_{j \rightarrow i} (t)$.
%
%
%
We skip zero-intensity windows whenever reaching $\lambda_{j \rightarrow i}(t) = 0$ during thinning, which is sound by viewing $N_{j \rightarrow i}(t)$ itself as a superposition of counting processes, one for each interval of non-zero intensity, and skipping their initial zero-rate periods by the memoryless property.\footnote{If $T \sim \text{Expo}(\lambda)$, then $P(T \geq t + s ~|~ T \geq s) = P(T \geq t)$.}
If $j$ recovers, i.e. $R_j(t) = 1$, then $\lambda_{j \rightarrow i}(t)$ in (\ref{eq:app-individual-exposure}) is dynamically set to 0 at the time of recovery because $K_{j,k}(t) = 0$.
By the principle of thinning, all exposure events caused by $j$ beyond this point, sampled back when $j$ got infectious, are discarded on the fly, \ie, when they get popped from $Q$.

Combining the above, we arrive at an efficient sampling procedure for the epidemiological model SDEs using a single priority queue $Q$, which is formally defined in Algorithms~\ref{alg:sampler-checkins} and~\ref{alg:individual-exposure} of  Appendix~\ref{app:sampling-algo}.
In this context, we note that interventions like social distancing or hygienic measures always \textit{reduce} the rates $\lambda_{j \rightarrow i}(t)$ and can thus likewise be implemented using thinning, \ie, rejecting the affected exposure events with some probability.

Our sampling procedure is a Las Vegas algorithm, \ie, its runtime is a random variable but its output, that is, the sampled trajectory of state variables, is always faithful. 
This is because the number of state transition events we sample and subsequently process depends on the number of infectious individuals, which is itself a random variable.
The following proposition bounds the expected runtime of our sampling procedure under some mild technical assumptions on the mobility traces $P_{i,k}(t)$.
A proof is given in Appendix~\ref{app:sampling-algo}.

\medskip 

{\em
\smallskip\noindent\textbf{Proposition 1.}~~~
    Assume that any given individual $i$~$ \in$~$\Vcal$ makes $O(\tmax)$ visits to sites $\Scal$, the mobility model is {\em sparse}, \ie, every individual $i \in \Vcal$ has $O(1)$ unique contact persons%
    \footnote{
    This implies that the individuals $\Vcal$ and sites $\Scal$ are not considered independently.
    Formally stated in terms of the mobility model, we assume $\sum_{k \in \Scal} P_{i,k} = O(\tmax)$ and $\sum_{j \in \Vcal} \sum_{k \in \Scal} P_{i,k}P_{j,k} = O(1)$.
    },
    and there are no containment measures.
    Then, the {\em expected} runtime of our sampling procedure for generating a trajectory of the epidemiological states $\mathbb{S}(t)$ over a time horizon $[0, \tmax)$ is given by
    \begin{align*}
        O \left(
         |\Vcal| \left (
            \tmax \log (\tmax |\Vcal|) 
            + \tfrac{1}{q} \tmax  \log (\tmax)
            \right )
        \right),
    \end{align*}
    where $q \in (0,1)$ is a constant known a priori that depends on the parameters of the epidemiological model.
     
}

\bigskip

\smallskip\noindent The above result implies that, if the number of sites individuals visit increases linearly with time and the number of unique contact persons is constant, then the expected runtime of our sampling procedure is {\em quasilinear} in the number of individuals $|\Vcal|$ and the length of the sampled trajectory $\tmax$. 
Moreover, it is worth pointing out that generating random rollouts of the model can be embarrassingly parallelized.
Finally, in our experiments, we have empirically found that our sampling procedure scales to regions of more than one hundred thousand individuals $\Vcal$ with around one thousand sites $\Scal$.

\subsection{Parameter Estimation}
\label{sec:sampling-inference}

Building on the sampling algorithm, 
we can estimate the unspecified epidemiological parameters $\theta = \{\beta_k, \xi\}$, 
\ie, the transmission rate of individuals at sites and in their households, in a given epidemiological scenario.
More specifically, provided
a set of initial conditions $\mathbb{S}(0)$,
testing policy $\pi_{\text{test}}$,
a priori fixed mobility traces $P_{i,k}(t)$,
and fixed parameters of the processes not concerning exposure, \ie, excluding $\{ N_i(t) \}_{i \in \Vcal}$,
we find the parameters $\theta$ that provide the best fit to the observed COVID-19 cases in a given region.
To this end, we view the model simulation as a black box and apply Bayesian optimization (BO), which amounts to iteratively building a surrogate model of our objective and evaluating at promising parameter settings~\cite{brochu2010tutorial}.

Following the standard BO paradigm, we interpret the \emph{expected number of positive cases at time $t$} in our model as a black box function $g_{t} (\theta)$ where
\begin{align}\label{eq:inference-simulator}
\displaystyle 
g_{t} (\theta)
&:= 
\mathbb{E}_{\Tcal \sim \theta} \Big [
\sum_{T_i^{+} \in \Tcal } 
T_i^{+}(t)
\Big]
\end{align}
The expectation in (\ref{eq:inference-simulator}) is defined over realizations of the testing state variables $\{T_i^{+}(t)\} =: \Tcal \sim \theta$ of the model with exposure parameters $\theta$.
In practice, $g_t(\theta)$ is only observed via noisy evaluations at different values of $\theta$ since the expectation is approximated using a Monte Carlo estimate of $J$ random simulations.
$\Tcal$ is stochastic not only due to the counting processes,
but in absence of real mobility traces also due to random seeds $\mathbb{S}(0)$ and synthetic $P_{i,k}(t)$, independently simulated for each rollout.
The objective we aim to minimize is the \emph{mean daily squared error} of cumulative positive cases between the model predictions and the real observed COVID-19 cases of the region.
This allows us to form a link between the spatiotemporal states of each individual in the model and aggregate longitudinal case data.
The squared error has previously been considered in parameter estimation for black-box models~\cite{astudillo2019bayesian} and in the context of COVID-19 research~\cite{chang2021mobility}.
Let $c^{\mathrm{true}}_{t}$ be the cumulative number of real COVID-19 cases at the end of day $t$ as provided by the national authorities.
Then, our objective $f$ to be minimized is a composition of the squared error score and per-day black-box functions $g_t(\theta)$ averaged over a time of $t_{\max}$ days:
\begin{align}\label{eq:inference-objective}
\displaystyle f (\theta)
&=
\frac{1}{T} \sum_{t=1}^{t_{\text{max}}}
\Big ( 
c^{\mathrm{true}}_{t} - g_{t}(\theta)
\Big )^2
\end{align}
The compositionality of $f$ can allow for greater sample efficiency~\cite{astudillo2019bayesian,bal2019botorch}, in particular when estimating additional parameters.
However, when only estimating $\theta = \{\{\beta_k\}, \xi\}$ and $\beta_k$ held constant across sites, we found it to be favorable for the BO surrogate model to directly learn $f(\theta)$, as opposed to the daily $g_t(\theta)$, as the black-box function.
We use the knowledge gradient acquisition function \cite{wu2016knowledge}  to navigate parameter proposals,
which often shows favorable performance in noisy settings~\cite{frazier2018tutorial,bal2019botorch}. 
Combining the above with the default BO procedure, our resulting parameter estimation algorithm is summarized in Algorithm~\ref{alg:estimation}.

\begin{algorithm}[t]
  \footnotesize
  \renewcommand{\algorithmicrequire}{\textbf{Input:}}
  \caption{Parameter estimation using Bayesian optimization} \label{alg:estimation}
    \begin{algorithmic}[1]
    \Require Black-box simulator ${g}_t(\theta)$,  parameter domain $\dom(\theta)$, time horizon $t_{\text{max}}$, case data $c_{t}^\mathrm{true}$, hyperparameters $J$, $M$, $N$
    
    \State $s (\mathbf{x}) := - \sum_{t=1}^{t_{\mathrm{max}}} (c^{\mathrm{true}}_{t} - \mathbf{x}_{t})^2$
    \State $\theta_{1:M} \gets $ first $M$ quasi-random settings
    \State $\Dcal \gets \emptyset$
    \For{$i \in [M]$} \Comment{Quasi-random initialization}
        \State Obtain daily sim. result  ${g}_t(\theta_i)$ from $J$ random roll-outs
        \State $\Dcal \gets \Dcal \cup \{ (\theta_i, {g}_t(\theta_i) )  \}$
        
    \EndFor 
    \While{$|\Dcal| \leq N$} \Comment{Bayesian Optimization}
        \State $p( {g}_t(\theta)) \;|\; \Dcal) \gets ~\text{GaussianProcessPosterior}(\Dcal) $
        \State $\theta^* \gets \arg \max_{\theta' \in \dom(\theta)} \text{KnowledgeGradient}(p, \theta' )$~\cite{wu2016knowledge}
        \State  Obtain daily sim. result  ${g}_t(\theta^*)$ from $J$ random roll-outs
        \State $\Dcal \gets \Dcal \cup \{ (\theta^*, {g}_t (\theta^* )) \}$
    \EndWhile 
    \State \Return $\arg \max_{(\theta, {g}_{1:t_{\mathrm{max}}}(\theta)) \in \Dcal}~ s({g}_{1:t_{\mathrm{max}}}(\theta))$
    \end{algorithmic}
\end{algorithm}

\section{A Case Study of Bern, Switzerland}
\label{sec:results}
In Sections \ref{sec:model} and \ref{sec:sampling-and-estimation}, we introduced a framework for epidemiological modeling and transmission parameter estimation in mobility models of any region of interest.
In summary, the application of our model presupposes 
\textit{(i)} a set of mobility traces $P_{i,k}(t)$ from synthetic mobility simulations or real-word data of the region, 
\textit{(ii)} the time distributions for disease progression after infection,
\textit{(iii)} initial conditions $\mathbb{S}(0)$ or assumptions about influx of infected individuals, and
\textit{(iv)} a testing policy $\pi_\mathrm{test}(t)$.

In the following, we showcase the flexibility of our framework in a case study of the city of Bern, Switzerland, and analyze the overdispersion of secondary infections as well as the course of the COVID-19 epidemic under various, fine-grained containment measures. 
We present supplementary results for additional regions of Germany and Switzerland in Appendix \ref{app:additional-results}.
More generally, the progression we follow in Sections \ref{sec:exp-setup} and \ref{ssec:results-estimation} can be viewed as step-by-step instructions to configure and calibrate the model to any desired region or disease variant.

\subsection{Experimental Setup} \label{sec:exp-setup}

\paragraph{Mobility traces}
We leverage fine-grained demographic data and open-source site locations to build a mobility model for Bern, Switzerland, that contains $|\Vcal|$~$=$~133,790 individual inhabitants visiting $|\Scal|$~$=$~2,174 real points of interest. 
The individuals $\Vcal$ belong to one of nine age groups according to the real demographics of the region.
They are placed in households of up to five people according to their age and reported household structure in Switzerland~\cite{household-structure-switzerland}.
The households themselves are located across the spatial expansion of the city using high-resolution population density data provided by \textit{Facebook Data for Good}~\cite{data-for-good}.
To obtain relevant site locations in the regions of interest, we use geolocation data 
provided by \emph{OpenStreetMap}~\cite{open-street-map}.
Specifically, we retrieve the location of all sites $\Scal$ in five site categories: \emph{education} 
(schools, universities, research institutes), \emph{social} (restaurants, caf\'es, bars), \emph{transportation} 
(bus stops), \emph{work} (offices, shops), and \emph{groceries} (supermarkets, convenience stores).
The sites $\Scal$ are visualized in Figure \ref{fig:maps-main}.

\looseness - 1 Since real check-in traces $P_{i,k}(t)$ of the population of Bern are not publicly available, we simulate synthetic mobility traces from the model in Section \ref{ssec:mobility} under the assumption of the gravity model~\cite{zipf1946p}.
In particular, we assume that each individual $i \in \Vcal$ visits only a constrained set of unique sites $\Scal_i \subset \Scal$, which 
are selected with probability inversely proportional to the squared distance from their homes.
This reflects the fact that individuals typically study or work at only one place, form habits regarding the public transportation they use, and social places or supermarkets they visit.
We set the check-in rate $\lambda_{i, k}(t)$ of Section \ref{ssec:mobility} to a constant value that depends on the individual's age group and site type; see Table~\ref{tab:mobrate} in the Appendix.
The mean duration $1/v_k$ at sites of type education and work, social, transportation, and groceries are fixed to 120, 90, 12 and 30 minutes, respectively. We sometimes set these times to lower values than one would expect because individuals are neither exposed to all others at a site nor continuously exposed during their visit.

\paragraph{Disease parameters}
As described in Section \ref{ssec:epidemiological-model},
we fix the parameters for disease progression after exposure based on recent estimates of the COVID-19 literature. The values we use are summarized in Table~\ref{tab:parameters} of the appendix.
We set mortality and hospitalization rates per age group using COVID-19 case data of the county-level administrative region \cite{covid-data-switzerland} and previous studies~\cite{ferguson2020impact}.

\paragraph{Testing}
To abstract away from testing criteria implemented in different regions, 
we assume that only true symptomatic individuals are registered for testing and that tests have perfect accuracy.
We set the reporting delay $\Delta_{\text{test}}$ to 30 minutes, accounting for the now frequently available\footnote{During parameter estimation, $\Delta_{\text{test}}$ is set to 48 hours to account for the test delay early in the pandemic.} rapid tests~\cite{rapid-tests}, and assume that there is sufficient capacity to test all selected individuals.
Moreover, positively tested individuals and their household members are quarantined for 14 days in isolation from each other.

\subsection{Model Fit and Parameter Estimation} \label{ssec:results-estimation}
To estimate the transmission rate at sites $\beta$ and in households $\xi$, we consider the time horizon from early March 2020 until May 2020 since it includes both times before and during governmental interventions in Switzerland, which occurred largely from March 16, 2020 to May 10, 2020 \cite{measures-switzerland}.
We use COVID-19 case data of the county-level administrative region \cite{covid-data-switzerland} to define the objective (\ref{eq:inference-objective}) and run the procedure described in Section~\ref{sec:sampling-inference} for $N \geq 100$ steps with $M=10$ initial quasi-random settings and $J = 200$ rollouts.
During governmental interventions, the check-in frequencies of individuals at sites in the mobility model are reduced as estimated by Google mobility data in the region~\cite{google-mobility-report}, and education and social sites are closed, \ie, not visited at all.
In the model simulations used for estimation, each of the $J$ realizations is randomized across realizations of the synthetic mobility traces and infection seeds. 
For parameter estimation, the initially exposed and infectious individuals are heuristically selected as described in Appendix \ref{app:state-initialization} based on knowledge about the case numbers at the start date of the estimation period.

\begin{figure}[!t]
    \centering
    \begin{tabular}{*{2}{b{0.5\textwidth-2\tabcolsep}}}
    \centering
    \includegraphics[width=0.85\linewidth,valign=c]{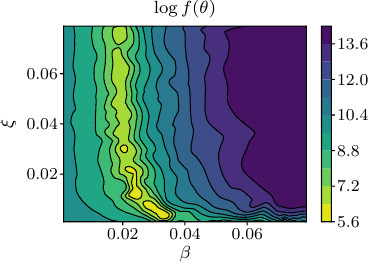}
    &
    \centering
    \hspace*{-30pt}
    \includegraphics[width=0.70\linewidth,valign=c]{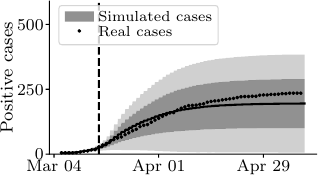}
    \end{tabular}
    \caption{{\bf Transmission parameter estimation for the model of Bern, Switzerland}. The left plot shows a contour plot of the log objective in (\ref{eq:inference-objective}) as a function of the transmission rate at sites $\beta$ and in households $\xi$. 
    The estimated parameters $\beta$~$=$~0.0337 and $\xi$~$=$~0.0038 lie in the identifiable optimal region.
    The right plot shows the predicted and real cumulative cases under the estimated parameters. The ``lockdown'' measures in Switzerland were implemented on March 16, 2020 as indicated by the dashed vertical line.
    The solid line indicates the mean, the shaded areas one and two standard deviations across 100 random simulations, respectively.
    }
    \label{fig:estimation}
\end{figure}

Figure \ref{fig:estimation} visualizes both the objective values $f(\thetab)$ obtained at various settings for the transmission rates $\beta$ and $\xi$ as well as the model predictions for the cumulative cases during the time window of parameter estimation.
The contour plot indicates that there is a single and identifiable optimal parameter regime, whose optimal values were estimated as $\beta$~$=$~0.0337 and $\xi$~$=$~0.0038.
Furthermore, we find that the simulations using the estimated parameters are able to accurately match the observed longitudinal trend of cases during the estimation period early in the epidemic.
Beyond the model of Bern, Switzerland, Appendix~\ref{app:additional-results} provides a collection of parameter estimation results for four additional regional models of other urban and rural regions in Switzerland and Germany~\cite{covid-data-germany,measures-germany-start, measures-germany-end}. 
These supplementary findings confirm a similarly identifiable optimal parameter regime and demonstrate that both the epidemiological model and the transmission parameter estimation procedure are robustly applicable to other regional mobility models.

\medskip 

In the remainder of this section, we use the estimated transmission parameters for the model of Bern in all of our experiments. 
We first empirically study the degree of overdispersion in the number of secondary infections caused by infectious individuals under our mobility and fitted transmission model.
We then use our framework to quantify the effects of a range of containment measures. 
To create a general epidemiological scenario, we assume a small but continual influx of five untraceable exogenous exposures per 100,000 inhabitants and per week and simulate the model state variables over a period of four months.

\subsection{Overdispersion of Secondary Infections}
As argued in Section \ref{sec:introduction}, existing epidemiological models have predominantly built on homogeneous Poisson transmission dynamics that fail to capture the overdispersion of secondary infections observed for COVID-19.
In addition, they do not explicitly model visits to sites where exposures occur. 
As a result, these models have been of little use for studying and predicting where and when infection hotspots are most likely 
to occur~\cite{zhang2020evaluating, frieden2020identifying,cevik2020sars,althouse2020stochasticity}.

In contrast to previous work, we find that {\em overdispersion of the distribution of secondary infections emerges naturally under our model}. 
Using the previously specified and estimated model parameters, we simulate the spread of COVID-19 under no containment measures other than the testing of symptomatic and isolation of positively tested individuals.
During these simulations, we count the number of secondary infections caused by individuals that got infectious during a 7-day window after 1 month of the model simulations.
Using two goodness-of-fit tests for the Poisson distribution, the Chi-squared ($\chi^2$) and variance tests (VT)~\cite{cochran1954some,fisher1992statistical}, we are able to reject the null hypothesis that the distribution 
of secondary infections, both overall and when stratified per visit, follows a Poisson distribution.
In particular, for both distributions of secondary infections, we obtain $\smash{p_{\chi^2}}$~$<$~$10^{-8}$ and $\smash{p_{\text{VT}}}$~$<$~$10^{-8}$.
With sample variance generally significantly exceeding the sample mean, both ways of counting the number of secondary infections naturally exhibit a higher variance than expected under the Poisson assumption and are thus \emph{overdispersed}.

\begin{figure}[!t]
    \centering
    \subfloat[Number of Secondary Exposures]{
        \centering
         \includegraphics[width=0.45\linewidth,valign=c]{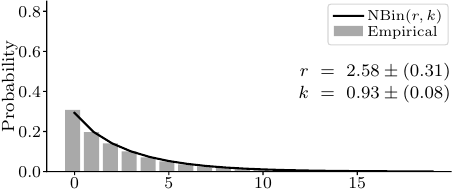}
        }
    \hspace{10pt}
    \subfloat[Number of Secondary Exposures During a Site Visit]{
        \centering
         \includegraphics[width=0.45\linewidth,valign=c]{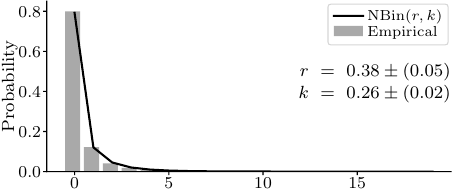}
        }
    \caption{
    {\bf Overdispersion of Secondary Infections.}
    Panels (a) and (b) show a generalized negative binomial model fitted by maximum likelihood estimation to the secondary infections caused by an infectious individual, overall and stratified by site visits, respectively, and averaged over 100 random realizations.
    The secondary infections are counted for individuals that got infectious during a 7-day window after 1 month of the simulation under no containment measures other than the testing of symptomatic and isolation of positively tested individuals. 
    }
    \label{fig:dispersion}
\end{figure}

To measure the degree of overdispersion, we follow recent work in the context of COVID-19 \cite{athreya2020effective,endo2020estimating} and fit a generalized negative binomial distribution $\text{NBin}(r, k)$, an overdispersed generalization of the Poisson, where $r > 0$ is the mean or reproduction
number, and $k$ is the dispersion parameter.
Figure~\ref{fig:dispersion} summarizes the results.
Averaged over 100 random realizations, we find that the dispersion parameter $k < 1$ both overall and when stratified per 
visit ($k = 0.93 \pm (0.08)$ and $k = 0.26  \pm (0.02)$, respectively), evidence of substantial overdispersion~\cite{endo2020estimating}. 
We hypothesize that the higher overdispersion observed when aggregating per visit is a direct effect of the interaction between the stochastic check-in mobility model and our model of transmission at sites.

\subsection{Efficacy of containment measures}

Reducing contacts at public sites by restricting individual mobility has been one of the most prevalent measures to 
counteract the spread of COVID-19~\cite{islam2020physical}. 
Our modeling framework allows us to faithfully study how effective various variants of this approach are at, \eg, 
containing the disease, reducing peak hospitalizations, or changing the effective reproduction number $R_t$ over time.
Instead of restricting the mobility of the entire population or only vulnerable groups, previous work has, for instance, 
proposed to divide the population into two subgroups that get isolated on alternating days~\cite{karin2020adaptive, meidan2020alternating}. 
\begin{figure}
    \centering
    \subfloat[Infected over time]{
        \centering
        \includegraphics[width=\linewidth]{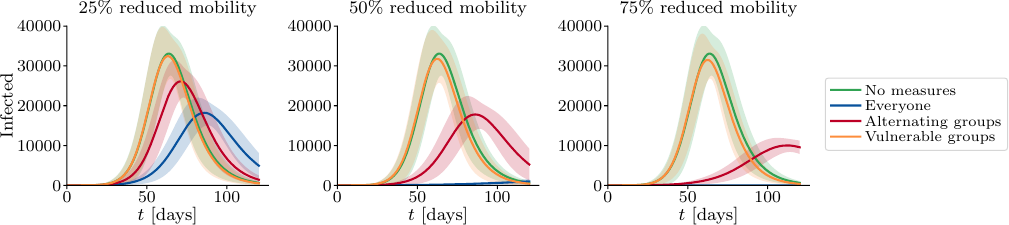}
        \label{fig:results-mobility-restrictions-over-time}
        }

    \subfloat[Effective reproduction number]{
        \centering
        \includegraphics[width=0.47\linewidth]{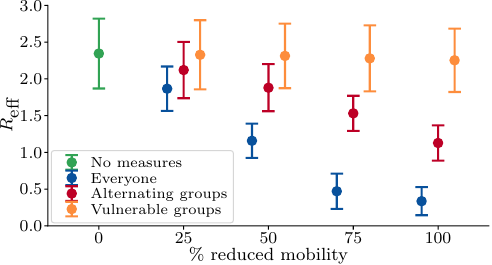}
        \label{fig:results-mobility-restrictions-reff}
        }
    \hspace{1em}
    \subfloat[Reduction of peak hospitalizations]{
        \centering
        \includegraphics[width=0.47\linewidth]{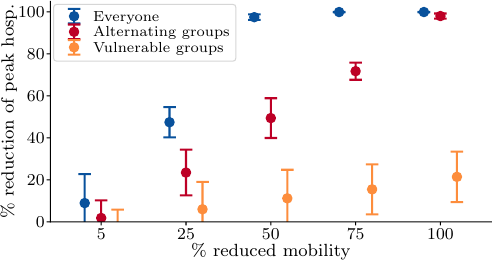}
        \label{fig:results-mobility-restrictions-hosp}
        }
        
    \caption{{\bf Mobility restrictions.} Individuals of certain groups reduce their mobility, i.e., their frequency of visits to sites by a certain proportion.
    Panel~a) shows the number of infected over time for different levels of mobility restrictions. 
    Panel~b) shows the effective reproduction number during the phase of exponential growth of the number of infected. 
    Panel~c) shows the reduction of peak hospitalizations compared to a scenario without mobility restrictions. 
    Points and lines represent the mean over 100 rollouts of the simulation. Error bars and shaded regions correspond to plus and 
    minus one and two standard deviations respectively.}
    \label{fig:results-mobility-restrictions}
\end{figure}

Figure \ref{fig:results-mobility-restrictions} shows a comparative analysis of three of these variants: restricting the 
mobility of everyone, only vulnerable groups, or one of two random subgroups on alternating days. 
The measures are implemented as described in  Sections \ref{ssec:testing-model}-\ref{ssec:containment-measures}.
For each variant, we consider different levels of mobility restriction where individual check-in activity at sites in the mobility 
model is reduced by between 5\% and 75\%.
In our simulations, the vulnerable groups are defined as individuals older than 60 years, who typically suffer more complications from COVID-19~\cite{covid-data-germany,covid-data-switzerland}.
Our findings highlight the fact that the efficacy of each policy strongly depends on the degree to which individual movement activity is reduced. 
While restricting the mobility of everyone is overall clearly most effective, our findings suggest that isolating (i.e., reducing the mobility by 100\%) one of two subgroups on alternating days can reduce the effective reproduction number, averaged over the phase of 
exponential case growth, and peak hospitalizations as much as reducing the mobility of everyone by 50\%.
Moreover, our results also suggest that the morally debatable strategy of quarantining only vulnerable groups does not live up 
to its expectation of reducing peak hospitalizations significantly.
\begin{figure}
    \centering
    \subfloat[Reduction of infections]{
        \centering
        \includegraphics[width=0.47\linewidth]{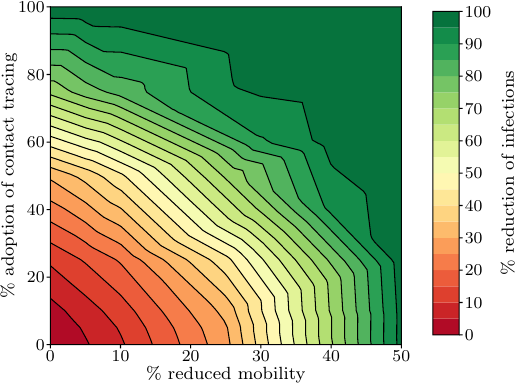}
        \label{fig:results-tracing-infections}
        }
    \hspace{1em}
    \subfloat[Reproduction number at $10$\% reduced activity]{
    \centering
    \includegraphics[width=0.44\linewidth]{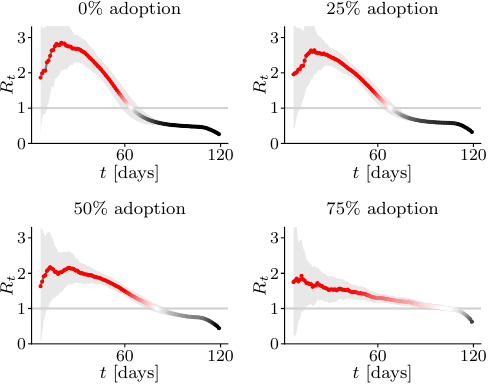}
    \label{fig:results-tracing-reff}
        }
        
    \subfloat[Infected over time]{
        \centering
        \includegraphics[width=\linewidth]{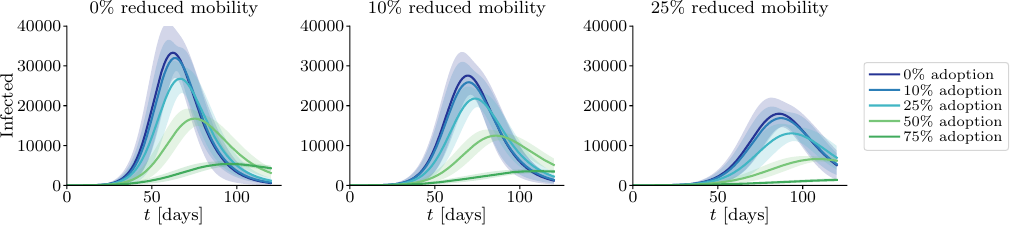}
        \label{fig:results-tracing-over-time}
        }
    \caption{{\bf Contact tracing.} A certain proportion of the population has adopted a digital contact tracing system and, in addition, everyone reduces their visit frequency to sites by a certain proportion. 
    Quarantine measures are imposed on traceable individuals who have been in contact with positively tested individuals for a minimum of $15$ minutes.
    Panel~a) shows the reduction of the cumulative number of infections compared to a scenario without interventional measures for different adoption levels of the tracing technology and various degrees of mobility restrictions. 
    Panel~b) shows the reproduction number over time for different adoption levels at a fixed mobility reduction of $10$\%. 
    Panel~c) shows the number of infected individuals over time for different scenarios.
    Lines and shaded regions represent the mean and two standard deviations computed over 100 rollouts of the simulation respectively.}
    \label{fig:results-tracing}
\end{figure}

\medskip

Orthogonal to various strategies that aim at reducing the number of contacts, the promise of digital contact tracing has been to achieve fine-grained epidemic control without severe societal or economical restrictions~\cite{ferretti2020quantifying}.
In this section, whenever an individual is tested positive, we use contact tracing to identify all of their contacts in the $10$ days leading up to the test result (see Section~\ref{ssec:containment-measures}).
If a given contact was longer than 15 minutes---the time threshold used by the national COVID-19 tracing apps in, \eg, France, Germany, Switzerland, and the United Kingdom~\cite{frenchcovidapp,germancovidapp,ukcovidapp,swisscovidapp}---the contact person is tested and isolated from everyone in the mobility model for $14$ days.

We analyze the effectiveness of digital contact tracing in combination with various degrees of mobility restrictions for the entire population at different digital tracing adoption levels.
The findings shown in Figure~\ref{fig:results-tracing-infections} illustrate that the adoption of the digital tracing system and the activity reduction due to social distancing have a complementary relationship in reducing the cumulative number of infections, as already argued by previous work~\cite{contreras2021challenges}.
Furthermore, the results suggest that, while contact tracing can provide a significant contribution to the mitigation of an epidemic, even at high adoption levels of $75$\%, it requires combination with activity reductions of $25$\% and above in order to achieve epidemic control ($R_t < 1$).
The effective reproduction number $R_t$ shown in Figure \ref{fig:results-tracing-reff} decreases over time at a constant adoption level due to the growing number of recovered individuals in the population.

\section{Related Work}
\label{sec:relatedwork}
Our work builds upon previous work on compartmental epidemiological modeling, human mobility models, and temporal point processes.
Most of the classical epidemiological literature has focused on developing population models~\cite{hethcote2000mathematics},
unable to capture heterogeneous transmission dynamics at the individual level.
More recently, there has been research on agent-based epidemic modeling
~\cite{ahn2013global,cator2012second,chakrabarti2008epidemic,van2011n,van2009virus}, 
also in the context of COVID-19~\cite{ferguson2020impact,moghadas2020projecting,wells2020impact,li2020substantial,diexpected,herbrich2020crisp,kucharski2020effectiveness}.
These models predominantly use multi-layer contact networks, discrete time, 
metapopulation, or Poisson transmission rate assumptions to characterize 
individual infections, rather than the frequency and duration of each 
individual'{}s visits to specific sites, as our model does.
Notable exceptions 
are by
Aleta et al.\ \cite{aleta2020modelling},
who use check-in data of real sites, yet only to configure the layers of a multi-layer 
contact network,
and Ferreti et al.\ \cite{ferretti2020quantifying}, who employ a time-varying transmission rate, but average over individuals who infect few or many others.
Chang et al.\ \cite{chang2021mobility} consider specific points of interest in US cities but only model transmission dynamics of metapopulations of up to 3,000 people rather than among single individuals.
Ultimately, none of the above models, including these three exceptions, can be faithfully used to characterize the dispersion of the number of individual infections during a visit, or to straightforwardly study the course of a disease under fine-grained intervention policies such as, \eg, contact tracing or testing.
As a result, these models have not been useful for studying conditions under which hotspots emerge~\cite{ frieden2020identifying,cevik2020sars}, analyzing measures to prevent SSEs~\cite{althouse2020stochasticity}, or predicting where infection hotspots are most likely to occur~\cite{zhang2020evaluating}.

The literature on human mobility models has a rich history, which has been extensively reviewed by Barcosa et al.\ \cite{barbosa2018human}.
In our experiments, the spatial distribution of visits of our \emph{event-based} ``check-in'' mobility model follows 
the gravity model~\cite{zipf1946p}.
Analogous to previous COVID-19 research, these visits are synthetically generated in each simulation~\cite{karin2020adaptive,duque2020timing,block2020social}.
However, our formulation is not restricted to this specific choice and one could think of 
designing event-based mobility models with a spatial distribution of visits following, \eg, the radiation model~\cite{simini2012universal} or population-weighted 
opportunities model~\cite{yan2014universal}. 
That said, the configuration of visit types, frequencies, and durations are specific to event-based models as ours, where the existing body of work on mobility models provides 
only very limited guidance~\cite{jankowiak2017uncovering,zarezade2018recurrent}.

Finally, there has been a flurry of work on temporal point process modeling in the machine
learning literature in recent years~\cite{zhou2013learning, xu2016learning, upadhyay2018deep,zuo2020transformer}. 
They have been particularly successful in predicting information propagation in social networks and the web, where they have achieved state of the art performances~\cite{du2016recurrent,farajtabar2017coevolve}. 
However, the development of compartmental epidemiological models based on temporal point processes has been lacking.

\section{Discussion}
\label{sec:discussion}
Motivated by multiple lines of evidence that strongly suggest for infection hotspots to play a key role in the transmission dynamics of COVID-19, we have introduced a spatiotemporal epidemic model that explicitly represents sites where infections occur and hotspots may emerge.
Through a case study that used fine-grained demographic data, site locations, mobility data as well as COVID-19 case data from Bern, Switzerland, we have demonstrated that our model can allow individuals and policy-makers to make more effective decisions concerning the implementation of containment measures, contact tracing, and testing---at the individual level and in the presence of overdispersion.
To facilitate this, we have released an easy-to-use implementation of the entire framework necessary to perform experiments for any desired region~\cite{implementation-model}.

While the purpose of this work does not lie in providing mechanistic forecasts, we have shown that an identifiable pair of only two fitted parameters, the transmission rates at sites and in households, provides reasonable predictiveness over our estimation window.
Importantly, we find that our epidemiological model empirically exhibits overdispersion in the number of secondary infections, which suggests that our formulation characterizes the transmission dynamics at infection hotspots---an epidemiological driver that effective containment measures would demand preventing \cite{althouse2020stochasticity,cevik2020sars}.
In this context, we do not intend to argue that our approach allows for a more accurate fit to aggregate case data than existing meta-population or network-based compartmental models.
Instead, our results in Section \ref{sec:results} demonstrate that we are able to formally model fine-grained interventions and perform analyses that would not be possible within the mathematical formulation used by existing meta-population models.

In this work, we have used fine-grained demographic data and site locations to configure our mobility model. 
However, if contact tracing data become accessible to researchers, we believe that the variance of our predictions 
could be lowered and that it would be possible to use our framework to identify areas with higher risk of infection 
in real time.
Beyond legal compliance and gaining societal acceptance, the use of epidemic models with high spatiotemporal resolution such as ours should respect each individual'{}s privacy.
It is hence important to highlight that, both during parameter estimation and contact tracing, we only need to compute the contact \emph{duration} of individuals with an infected person---the identity of the infected person is not required. As a result, there are reasons to believe that such computations can be made in a decentralized and privacy-preserving manner~\cite{nanni2020data, private-contact-tracing}.
Ultimately, although our model has greater resolution than many of those in use today, 
its predictions can only be faithfully considered when being aware of the high variance observed across random 
realizations.

\section*{Acknowledgements}
\label{sec:acknowledgements}
We thank the Robert-Koch-Institute, Open\-Street\-Maps, Google, and Facebook for providing data to make this work possible.
We thank Brian Karrer from Facebook for his insightful comments and suggestions regarding Bayesian optimization,
Kevin Murphy, Yusef Shafi and others from Google for helpful discussions,
and Yannik Schaelte for useful comments on a preliminary version of this work. 
We thank Cansu Culha and the Stanford Future Bay Initiative as well as Pavol Harar from the University of Vienna for working with us to improve our publicly available implementation.
This work was supported in part by SNSF under grant number 200021-182407 and the European Research Council (ERC) under the European Union'{}s Horizon 2020 research and innovation programme (grant agreement No. 945719).

\setlength{\bibsep}{3pt plus 1pt}
{
\hypersetup{urlcolor=gray}
\small
\bibliographystyle{naturemag}
\bibliography{refs}
}

\appendix
\newpage

\section{Household Exposures}
\label{app:household}

If information about households $\Hcal(i)$ that each individual $i \in \Vcal$ belongs to is available, one can account for exposures within households analogously to exposures at sites $\Scal$ by adding an additional rate $\lambda_{\Hcal(i)}(t)$ to the conditional intensity function $\lambda_i(t)$ of the exposure counting process $N_i(t)$:
\begin{equation}
    \lambda_{\Hcal(i)}(t) = S_i(t) \, \, \xi \, \sum_{j \in \Hcal(t) \backslash i} \int_{t-\delta}^{t} 
     \,K^\Hcal_{i,j}(\tau) \,
     \gamma e^{-\gamma(t-\tau)}d\tau
\end{equation}
where 
\begin{align}
\begin{split}
    K^\Hcal_{i,j}(\tau) = 
    &\Big(I^{s}_j(\tau) + I^{p}_j(\tau) + \mu I^{a}(\tau) \Big)  \prod_{k \in \Scal}  (1-P_{i,k}(\tau)) 
    (1-P_{j,k}(\tau)) 
\end{split}
\end{align}
where $\xi \geq 0$ is the base transmission rate within households. This intensity function models our assumption that individuals within a household are in contact as long as they are not visiting any site.

Exposure events caused by $\lambda_{\Hcal(i)}(t)$ can be sampled analogously to the principles for sampling exposure times introduced in Section~\ref{sec:sampling}. Their superposition with exposures at sites is handled by the priority queue.

\section{Empirical Probability of Exposure}
\label{app:exposure-probability}

The exposure risk of others caused by an infectious individual can be computed under our model and empirically approximated using location or contact data, \eg, from (manual) contact tracing.
Specifically, the probability of exposure $\hat{p}_{i \leftarrow j}(t_0, t_f)$ during a time window $[t_0, t_f]$ associated with $j$ in the process $N_i(t)$ is given by:
\begin{equation}\label{eq:advanced-contact-tracing}
\hat{p}_{j \leftarrow i}(t_0, t_f) = 1 -
\exp \left( -  K^{\text{risk}}_{j,i}(t_0, t_f) \right) 
\end{equation}
with
\begin{align}
\hspace*{-3pt} K^{\text{risk}}_{j,i}(t_0, t_f) =  \sum_{k \in \Scal} \beta_k \hspace*{-3pt}  \int_{t_0}^{t_f} \hspace*{-4pt}   P_{i, k}(t') \hspace*{-4pt}  \int_{t'-\delta}^{t'} P_{j,k}(\tau)  \gamma e^{-\gamma(t'-\tau)}  d\tau dt'  
\end{align}
and follows from the survival probability in a temporal point process~\cite{de2019temporal}.
The estimated probability of exposure is conservatively high by assuming that all contacts are (pre-)symptomatic and not considering a possible scaling of $\mu$ for asymptomatic individuals. See  Section \ref{ssec:epidemiological-model}.

\section{State variable initialization}
\label{app:state-initialization}

During the parameter estimation period, it is necessary to specify initial epidemiological conditions $\mathbb{S}(0)$ that are consistent with the COVID-19 case data used in the objective.
To this end, we set the number of initially symptomatic individuals $I^s_{\text{init}} = \sum_{i\in \Vcal} I^s(0)$ equal to the real observed COVID-19 cases in a region at the start date, or scaled proportionally to the population size in an administrative region, and set all to be positively tested.
Based on the above, we seed $I^a_{\text{init}} = \alpha_a / (1-\alpha_a) I^s_{\text{init}}$ individuals to be initially asymptomatic to obtain a proportion of recently estimated $\alpha_a = 0.4$ asymptomatic seeds~\cite{nishiura2020estimation,lavezzo2020suppression,ferretti2020quantifying}.
Assuming that infectious individuals have exposed $R_0$ others on average, we seed $E_{\text{init}} = R_0 (I^a_{\text{init}} + I^s_{\text{init}})$ initially exposed individuals, 
using recent estimates of the basic reproduction number of approximately $R_0 \approx 2.0$ \cite{tindale2020r,who-final-report,ferretti2020quantifying}.
In any simulation done for parameter estimation,
$E_i(0), I^a_i(0), I^s_i(0)$ are seeded uniformly at random following the above heuristic counts. 
Neither asymptomatic nor symptomatic seeds cause further exposures, and for simplicity no other states are initially seeded.

\newpage

\section{Sampling procedure} \label{app:sampling-algo}

\subsection{Algorithms}
\begin{algorithm}[!h]
  \small
  \renewcommand{\algorithmicrequire}{\textbf{Input:}}
  \caption{Sampling algorithm for generating a trajectory of the epidemiological states $\mathbb{S}(t)$\protect\footnotemark}
\label{alg:sampler-checkins}
    \begin{algorithmic}[1]
    \Require Initial conditions $\mathbb{S}(0)$, mobility traces $P_{i, k}(t)$, parameters $\gamma$, $\delta$, $\alpha_a$, $\alpha_b$, $\alpha_h$, $\mu$ and $\beta_k$, rates $\lambda_{(\cdot)}(t)$ of the counting processes 
   
    \State $\tnow \gets 0$, $S_i \gets 1$, $Q \gets$ priority queue processing in temporal order of events

    \For{all $i \in \mathcal{V}$ s.t. $S_i = 0$}\Comment{Initial conditions}
        \State  Push initial state transition $(0, \cdot, i, \varnothing, \varnothing)$ to $Q$ 
    \EndFor

    \While{$Q$ not empty}\Comment{Priority queue $Q$ contains $(\text{time}, \text{transition}, i, \text{infector}, \text{site})$ events}
    \vspace*{3pt}
    \State  $(\tnow, e, i, j, k) \gets$ pop earliest from $Q$ 
    %
    %
    \vspace*{3pt}
    \If{$e$ \textbf{is} $dE$ \textbf{and} $R_j(\tnow) = 0$ \textbf{and} $D_j(\tnow) = 0$ \textbf{and $S_i = 1$}} 
        \vspace*{2pt}
        \If{\Call{Interventions}{$i, j, k, \tnow$}}  \Comment{Reject event and re-sample arrival time of event?}
            \State Call Algorithm \ref{alg:individual-exposure} with $(P, j , i, \tnow, r= 1 - (1-\mu)I^{a}_j(\tnow) )$  
        \Else \Comment{Person $i$ exposed by infector $j$}
            \State  $E_i \gets 1$, $S_i \gets 0$, $\Delta_M \sim \text{Expo}(\lambda_M(\tnow))$ , $u \sim \text{Unif}(0, 1)$ 
        
	        \If{$u \leq \alpha_a$}
        	    \State Push $(\tnow + \Delta_M, dI^a, i, \varnothing, \varnothing)$ event to $Q$ 	\Else
		        \State Push $(\tnow + \Delta_M, dI^p, i, \varnothing, \varnothing)$ event to $Q$  
	        \EndIf
	    \EndIf
    %
    %
    \vspace*{3pt}
    \ElsIf{$e$ \textbf{is} $dI^p$} \Comment{Person $i$ presymptomatic}
    	\State  $I^p_i \gets 1$, $E_i \gets 0$, $\Delta_Z \sim \text{Expo}(\lambda_W(\tnow))$
		\State Push $(\tnow + \Delta_W, dI^s, i, \varnothing, \varnothing)$ event to $Q$  
		\For{$u$ such that $S_u = 1$}
		
		\State Call Algorithm \ref{alg:individual-exposure} with  arguments $(P, i, u, \tnow, r=1)$
		\EndFor
    %
    %
    \vspace*{3pt}
    \ElsIf{$e$ \textbf{is} $dI^s$} \Comment{Person $i$ symptomatic}
   
   	\State  $I^s_i \gets 1$, $I^p_i \gets 0$, $u, v \sim \text{Unif}(0, 1)$         
     \If{$u \leq \alpha_h$}
        \State $\Delta_{Y} \sim \text{Expo}(\lambda_{Y}(\tnow))$, Push $(\tnow + \Delta_Y, dH, i, \varnothing, \varnothing)$  to $Q$ 
     \EndIf
	 \If{$v \leq \alpha_b$}
	 	\State $\Delta_{Z} \sim \text{Expo}(\lambda_{Z}(\tnow))$, Push $(\tnow + \Delta_Z, dD, i, \varnothing, \varnothing)$  to $Q$ 	
	\Else
		\State $\Delta_{R} \sim \text{Expo}(\lambda_{R^s}(\tnow))$, Push $(\tnow + \Delta_{R}, dR, i, \varnothing, \varnothing)$  to $Q$  
	\EndIf

    %
    %
    \vspace*{3pt}
    \ElsIf{$e$ \textbf{is} $dI^a$} \Comment{Person $i$ asymptomatic}
    	\State  $I^a_i \gets 1$, $E_i \gets 0$, $\Delta_{R} \sim \text{Expo}(\lambda_{R^a}(\tnow))$
        \State Push $(\tnow + \Delta_R, dR, i, \varnothing)$ event to $Q$ 
	\For{$u$ such that $S_u = 1$}
		\State Call Algorithm \ref{alg:individual-exposure} with arguments $(P, i, u, \tnow, r=\mu)$
	\EndFor
	
    %
    %
    \ElsIf{$e$ \textbf{is} $dH$} \Comment{Person $i$ hospitalized}
    	\State  $H_i \gets 1$
    %
    %
    %
    \ElsIf{$e$ \textbf{is} $dR$} \Comment{Person $i$ resistant}
    	\State  $R_i \gets 1$, $I^a_i \gets 0$, $I^s_i \gets 0$, $H_i \gets 0$
    %
    %
    \ElsIf{$e$ \textbf{is} $dD$} \Comment{Person $i$ died}
    	\State  $D_i \gets 1$, $I^s_i \gets 0$, $H_i \gets 0$
    \EndIf
    \EndWhile
    \end{algorithmic}
\end{algorithm}

\footnotetext{For simplicity, we omit details about the procedure {\textproc{Interventions}$(i, j, k, t)$}, which applies thinning as explained in Section \ref{sec:sampling} for possible interventional measures.
Details can be found in our publicly available implementation~\cite{implementation-model}.}

\begin{algorithm}
  \small
  \renewcommand{\algorithmicrequire}{\textbf{Input:}}
  \caption{Push the next event of individual $i$ exposing individual $j$ in time window $[t, t_{\max})$ by considering only the contribution $\lambda_{i \rightarrow j}(t)$ in  (\ref{eq:app-individual-exposure})
   as described in Section~\ref{sec:sampling}.}\label{alg:individual-exposure}
    \begin{algorithmic}[1]
    \Require $P$, $i$, $j$, $t$, $r$
    
    \Procedure{InContact}{$u, v, \tau$} 
        \State \Return True \textbf{if} $\exists k \in \Scal$ s.t.  $P_{u,k}(\tau) = 1$   and  $\exists \tau' \in [\tau - \delta, \tau]$ s.t. $P_{v,k}(\tau') = 1$
        \Statex \hspace*{10pt} \textbf{else} \Return False
    \EndProcedure
    \Procedure{NextContact}{$u, v, \tau$}
        \State \Return $ \min_{\tau' > \tau} \tau'$ s.t. \Call{InContact}{$u, v, \tau'$}
    \EndProcedure
    \Procedure{WillBeInContact}{$u, v, \tau$}
        \State \Return True \textbf{if} $\exists \tau' \in [\tau, t_{\max}]$ s.t. \Call{InContact}{$u, v, \tau'$}$]$
        \Statex \hspace*{10pt} \textbf{else} \Return False
    \EndProcedure

    \State $\tau \gets t$ 
    \While{\Call{WillBeInContact}{$j, i, \tau$}} \Comment{Thinning loop}
        
        \State $b \gets $ \Call{InContact}{$j, i, \tau$}
        \If{\textbf{not} $b$} 
              \State $\tau \gets$ \Call{NextContact}{$j, i, \tau$} 
         \EndIf
            
        \State $\Delta_{E_j} \sim \text{Expo}\left ( \max_k\{\beta_k\} \, r \int_{\tau-\delta}^\tau \gamma e^{- \gamma (\tau - v)}  dv \right )$
        \State $\tau \gets \tau + \Delta_{E_j}$
        \If{\Call{InContact}{$j, i, \tau$}}
            \State $k \gets $ site of contact 
            \State $p \gets \left ( \beta_k \, r  \int_{\tau-\delta}^\tau \gamma e^{-\gamma (\tau - v)} P_{i, k} (v)dv \right ) \big /  \left ( \max_k\{\beta_k\}  \, r  \int_{\tau -\delta}^\tau \gamma e^{-\gamma (\tau - v)} dv \right )$ 
            \State $u \sim \text{Unif}(0, 1)$ 
            \If{$u \leq p$}\Comment{Accept/reject sampled event time}
                \State Push $(\tau, dE, j, i, k)$ event to $Q$
                \State \textbf{break}
        \EndIf
        \EndIf
    \EndWhile

    \end{algorithmic}
\end{algorithm}

\clearpage 

\subsection{Proof of Proposition 1}
The sampling algorithm and its subroutine are formally defined in Algorithms \ref{alg:sampler-checkins} and \ref{alg:individual-exposure}.
In the following, we assume that: 
(i) a given individual $i~$~$\in$~$\Vcal$ makes $O(\tmax)$ visits to sites $\Scal$ over the horizon $[0, \tmax)$;
(ii) the mobility model is {\em sparse}, \ie, every individual $i \in \Vcal$ has $O(1)$ unique  contact persons; and,
(iii) there are no containment measures.
This implies that there are a total number of $O(\tmax)$ contact windows of $i$ with all other individuals $j \in \Vcal$.
Following our implementation \citep{implementation-model}, we assume that the mobility traces $P_{i,k}(t)$ are given as an unsorted list of time intervals $[t_0, t_1]$, where each time interval indicates a visit of an individual $i \in \Vcal$ to a site $k \in \Scal$ during the simulated time period. 

\paragraph{Event queue}
In any possible trajectory of the epidemiological state variables, there is a constant number of events {\em not} concerning exposure that can be pushed to the event queue $Q$ per individual. 
This is because every individual transitions through at most a finite set of states.
In addition, since by assumption (ii) the mobility model is sparse, there is a constant number of exposure events caused by and thus pushed per individual. 
Thus, the overall number of events pushed to the event queue throughout the simulation is $O(|\Vcal|)$.
This is an upper bound on the size of the queue at any point in the simulation.
Using the standard heap implementation of a priority queue, pushing to and popping from the temporally-sorted event queue $Q$ hence incur cost $O(\log (|\Vcal|))$ in the worst case.

\paragraph{Preprocessing of contacts}
Sampling exposures caused by an infectious individual $i$ relies on querying the contacts with other individuals $j$ by checking their overlapping visits to sites $\Scal$.
To do this efficiently, the mobility traces  are preprocessed into efficient interval data structures called {\em interval trees}.%
\footnote{An interval tree containing $n$ intervals allows for $O(\log n)$ insertion time. 
Using binary search, retrieving the subset of stored intervals that intersect with a query interval $[t_0, t_1]$ takes time $O(m + \log n)$, where $m$ is the number of intersecting intervals.}
For this, we initialize two dictionaries that store two kinds of interval trees, visits by individuals and visits to sites, respectively.
Both dictionaries  are populated by iterating once over all $O(\tmax|\Vcal|)$ site visits in the simulated period.
For each visit, its time interval is inserted both into the tree of visits by the corresponding individual $i \in \Vcal$ as well as the tree associated to the site $k \in \Scal$.
Then, by assumption (i), the interval trees stratified by {\em individual} have size $O(\tmax)$ and intervals do not overlap by construction.
Moreover, the interval trees stratified by {\em site} contain $O(\tmax|\Vcal|)$ visits and, by assumption (ii), any interval overlaps with $O(1)$ others.
Thus, the total time incurred for constructing all visit interval trees is $O(\tmax|\Vcal| \log (\tmax|\Vcal|))$.

Using these two sets of {\em visit} interval trees, we build a collection of $O(1)$ {\em contact} interval trees for each individual $i \in \Vcal$.
These contain the contact windows from $i$ to each of its unique contact persons $j$.%
\footnote{We denote a contact as being {\em from $i$ to $j$} to be precise about non-contemporaneous infections (cf.\ Equation~(\ref{eq:intensity-y})).
There is an exposure-relevant contact from $i$ to $j$ if $i$ left a site less than $\delta$-time before $j$ arrived.} 
To generate the contact trees for $i$, we iterate over all $O(\tmax)$ visits of $i$. For each visit, we query the interval tree of the visited site in time $O(\log(\tmax |\Vcal|))$ to retrieve the $O(1)$ contact persons $j$ during that visit. Given this, we update the individual contact interval tree from $i$ to $j$ in time $O(\log (\tmax))$. 
Like individual visit traces, the contact intervals do not overlap by construction.
The overall preprocessing cost remains $O( \tmax|\Vcal| \log (\tmax|\Vcal|))$.

\paragraph{Handling events}
The backbone of the sampling procedure in Algorithm \ref{alg:sampler-checkins} consists of processing state transition events in temporal order.
All generic state transitions in the model, \ie, those {\em not} transitioning to an infectious state, consist of updating the correct indicator variables of the corresponding individual $i$ or discarding events that became invalid due to thinning in constant time.
In addition, we push the next state transition of $i$ to $Q$, which takes time $O(\log(|\Vcal|))$.
Since there are $O(|\Vcal|)$ generic events in the worst case, handling all of them takes an overall time of $O(|\Vcal| \log (|\Vcal|))$.

When an individual $i$ first transitions to an {\em infectious} state, \ie, the presymptomatic $I^p_i=1$ or asymptomatic $I^a_i=1$ state,
an additional time cost is incurred because we sample the times of the exposure events {\em caused} by $i$ to all of its unique contact persons $j$ in the future. 
This corresponds to calls of Algorithm~\ref{alg:individual-exposure}, where we continually iterate over all $O(\tmax)$ contact windows $i$ has with $j$ after some time $t$ {\em until the first valid exposure event is sampled}.
Specifically, we sample a next time $t$ as $t \gets t + \tau$ with $\tau \sim \text{Expo}(\lambda_{\max})$.
If $i$ is still in contact with $j$ at time $t$, and if the event is not rejected using thinning due to a lower site-specific exposure rate $\beta_k$ or asymptomatic infectiousness $\gamma$, 
the exposure time is valid and we push the event to $Q$ in time $O(\log (|\Vcal|))$.
Otherwise, we repeat.
Since there are at most $O(\tmax)$ contact windows of $i$ with $j$, 
each query to \textproc{InContact} as formalized in Algorithm~\ref{alg:individual-exposure} incurs time $O(\log (\tmax))$ using the interval tree.

Let $U$ be the random variable representing the runtime of Algorithm \ref{alg:individual-exposure} incurred by {\em one} contact window from $i$ to $j$.
In addition, let $q \in (0,1)$ be the probability that a given thinning sample gets accepted.
By the memoryless property of thinning, the expected value of $U$ is given by
\begin{align}
\begin{split}\label{eq:thinning-time-complexity}
    \EE[U] 
    &= O(\log (\tmax)) + q \, O(\log (|\Vcal|)) + (1-q) \EE[U] \\
    &= \sum_{n=0}^\infty (1-q)^n \big ( O(\log (\tmax)) + q \, O(\log (|\Vcal|)) \big)
    = O(\log (|\Vcal|)) + \tfrac{1}{q} O(\log (\tmax))
\end{split}
\end{align}
In the worst case, thinning is done for all $O(\tmax)$ contact windows from $i$ to $j$ until an exposure event time gets accepted.
Overall, Algorithm \ref{alg:individual-exposure} is called $O(1)$ times per individual.
Thus, the processing of state transitions to infectious states of all $O(|\Vcal|)$ infectious individuals incurs an additional overall cost of $O(|\Vcal| \tmax (\log (|\Vcal|) + \tfrac{1}{q} \tmax  \log (\tmax)))$.
This also accounts for the cost of sampling household exposures, which can be viewed as visits to an additional site with an additional set of $O(1)$ household contacts.
We note that $q$ is a constant that depends only on the exposure rate of the epidemiological model, and any lower bound thereof across sites and individuals suffices for Equation~(\ref{eq:thinning-time-complexity}).

\paragraph{Expected runtime}
Combining the preprocessing cost, the handling of all generic state transitions, and the handling of transitions to infectious states, Algorithm \ref{alg:sampler-checkins} has a total expected runtime of
\begin{align}
    O \left(
        |\Vcal| \left (
            \tmax \log (\tmax |\Vcal|) 
            + \tfrac{1}{q} \tmax \log (\tmax)
        \right )
    \right)
    \qed
\end{align}

\clearpage
\newpage

\section{Estimation Results for Additional Regional Models}
\label{app:additional-results}
Figure \ref{fig:estimation-appendix} summarizes the parameter estimation results for four additional regions in Germany and Switzerland: the cities of T{\"u}bingen and Kaiserslautern as well as the Canton of Jura and the district Rheingau-Taunus.
As for the model of Bern, the estimation procedure was executed as described in Section \ref{sec:exp-setup}.
Table \ref{tab:parameters} lists the estimated optimal parameters as well as additional details about each regional model.

\begin{figure}[!h]
    \centering
    \subfloat[T{\"u}bingen, Germany]{
        \begin{tabular}{cc}
        \centering
            \multirow{2}{*}[20pt]{
            \centering
                \includegraphics[width=0.55\linewidth,valign=c]{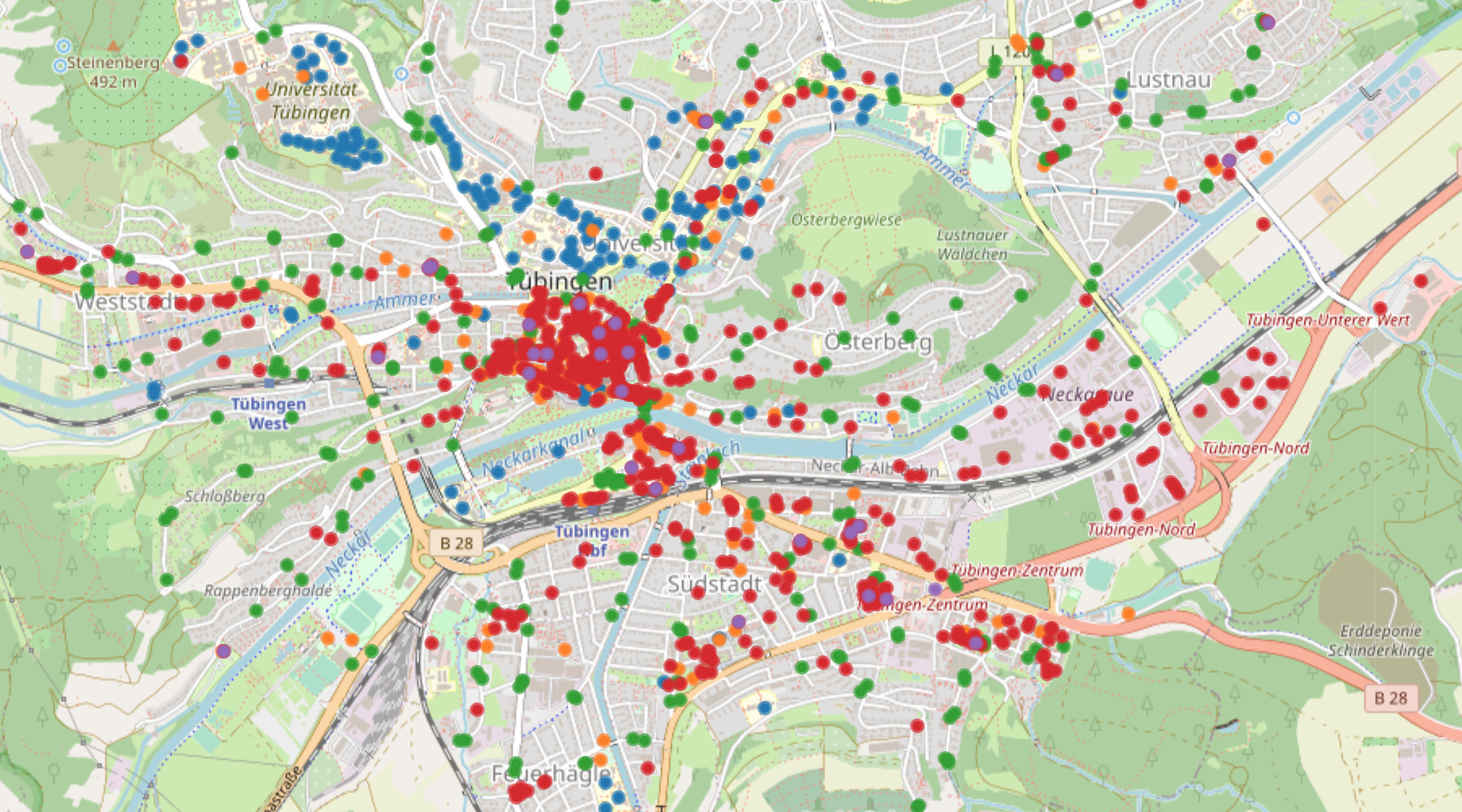}
            }
            &
            \includegraphics[width=0.38\linewidth,valign=c]{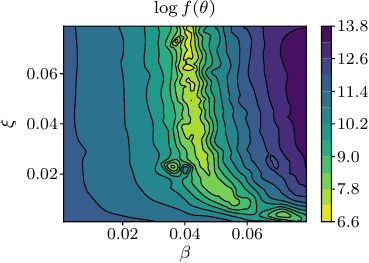}
            \\
            &
            \hspace*{-8pt}\includegraphics[width=0.33\linewidth,valign=c]{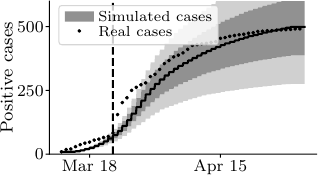}
        \end{tabular}
        \label{fig:estimation-appendix-TU}
    }\\
    \subfloat[Canton Jura, Switzerland]{
        \begin{tabular}{cc}
        \centering
            \multirow{2}{*}[20pt]{
            \centering
                \includegraphics[width=0.55\linewidth,valign=c]{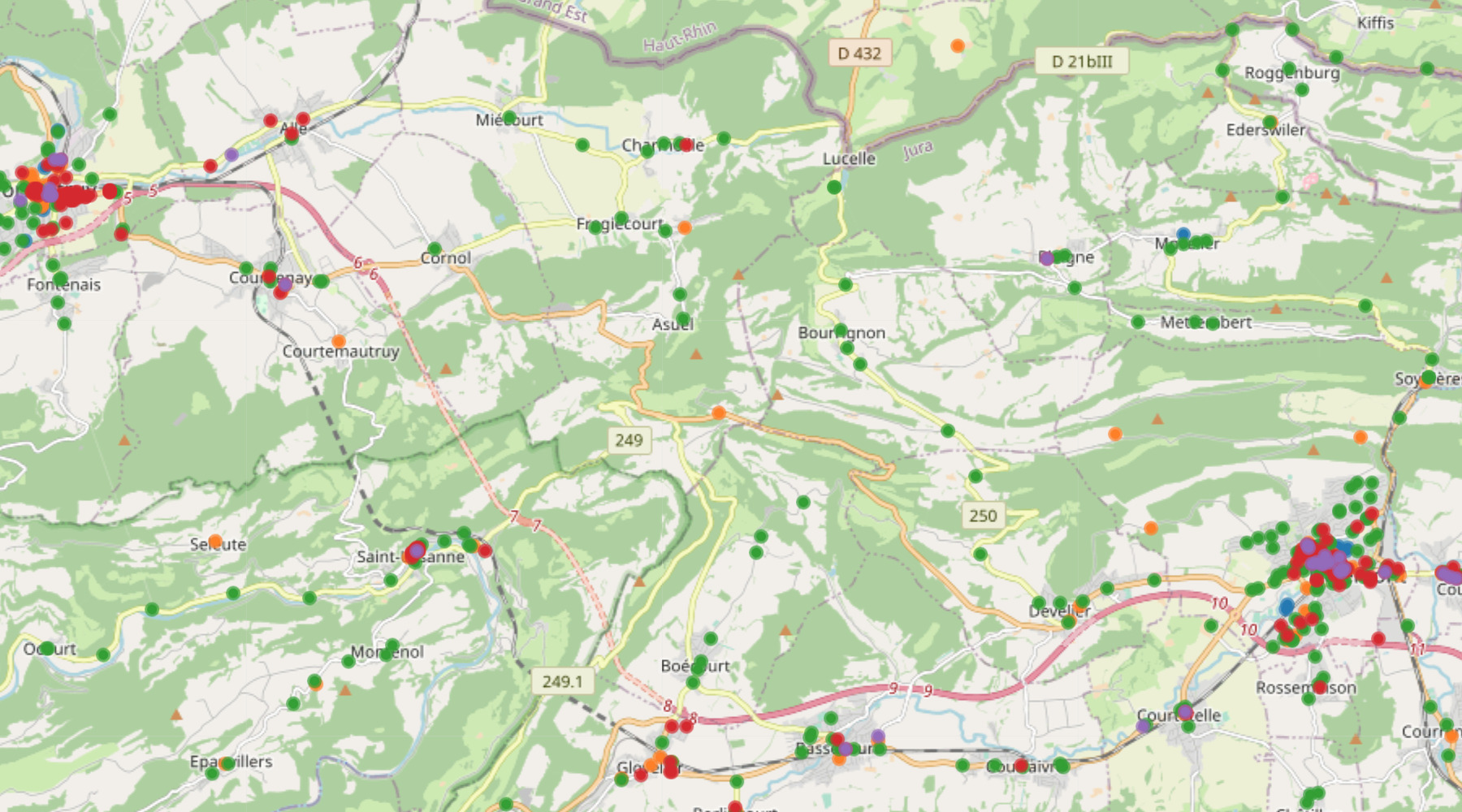}
            }
            &
            \includegraphics[width=0.38\linewidth,valign=c]{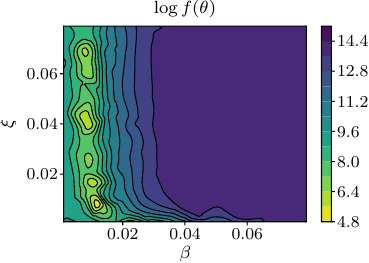}
            \\
            &
            \hspace*{-8pt}\includegraphics[width=0.33\linewidth,valign=c]{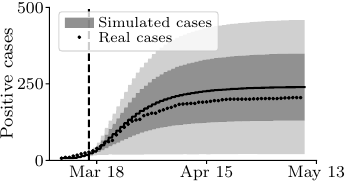}
        \end{tabular}
        \label{fig:estimation-appendix-JU}
    }
    \caption{\looseness-1{\bf Transmission parameter estimation for additional regional models in Germany and Switzerland.}
    In each panel, the left plot shows the sites of the mobility model with colors used as in Figure~\ref{fig:maps-main}.
    The right plots show the contour plot of the log objective in (\ref{eq:inference-objective}) as a function of the transmission rates $\beta$ and $\xi$ as well as the predicted and real cumulative cases under the estimated parameters, analogous to Figure~\ref{fig:estimation}.
    }
    \label{fig:estimation-appendix}
\end{figure}

\begin{figure}[!h]
    \addtocounter{figure}{-1}
    \setcounter{subfigure}{2}
    \centering
    \subfloat[Rheingau-Taunus-Kreis, Germany]{
        \begin{tabular}{cc}
        \centering
            \multirow{2}{*}[20pt]{
            \centering
                \includegraphics[width=0.55\linewidth,valign=c]{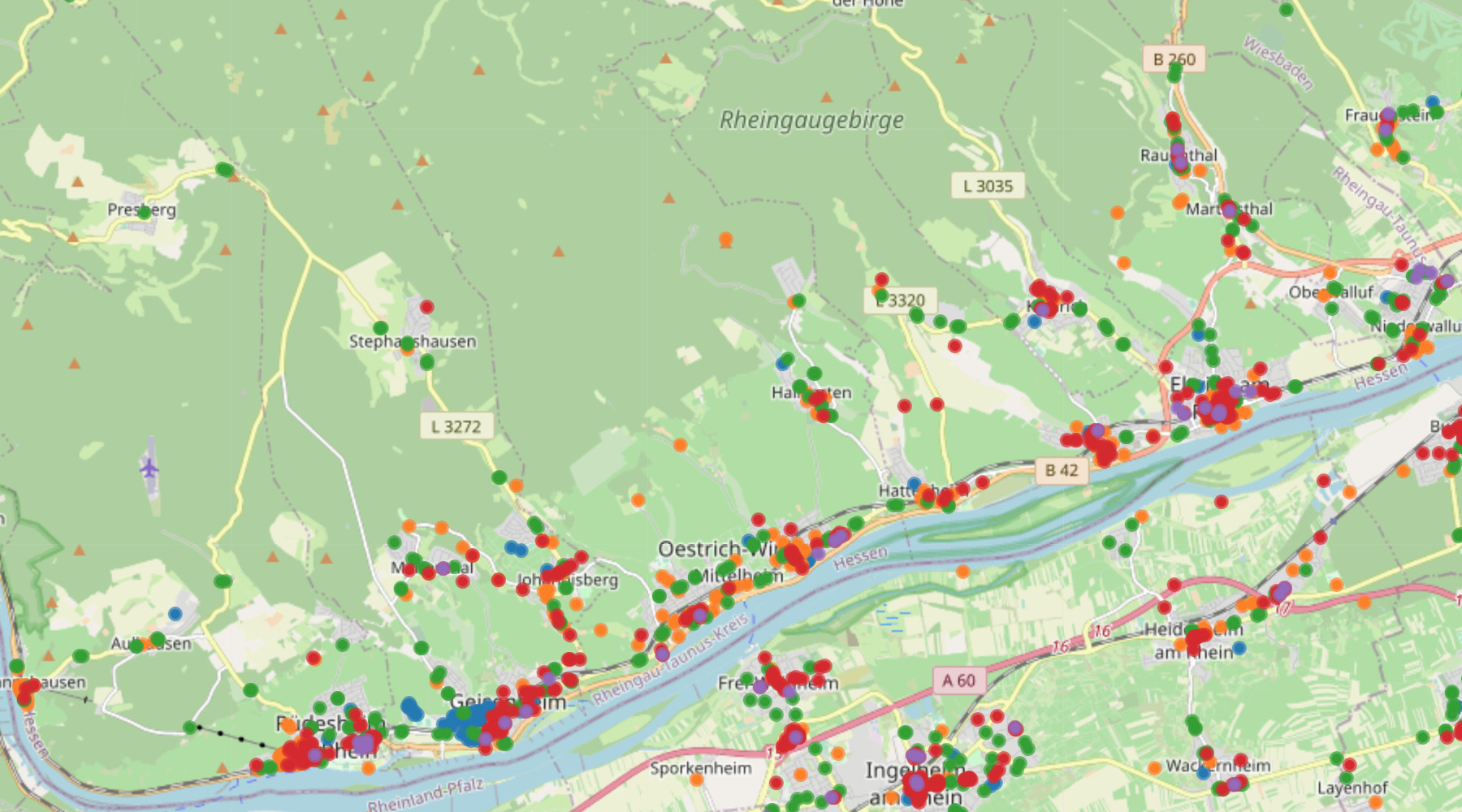}
            }
            &
            \includegraphics[width=0.38\linewidth,valign=c]{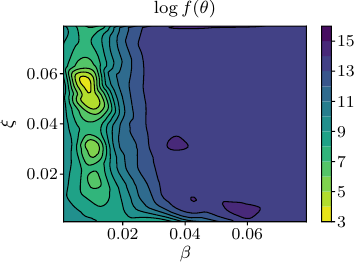}
            \\
            &
            \hspace*{-8pt}\includegraphics[width=0.33\linewidth,valign=c]{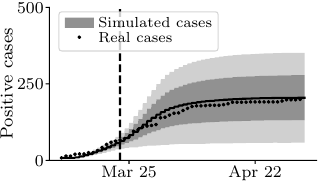}
        \end{tabular}
        \label{fig:estimation-appendix-RH}
    }\\
    \subfloat[Kaiserslautern, Germany]{
        \begin{tabular}{cc}
        \centering
            \multirow{2}{*}[20pt]{
            \centering
                \includegraphics[width=0.55\linewidth,valign=c]{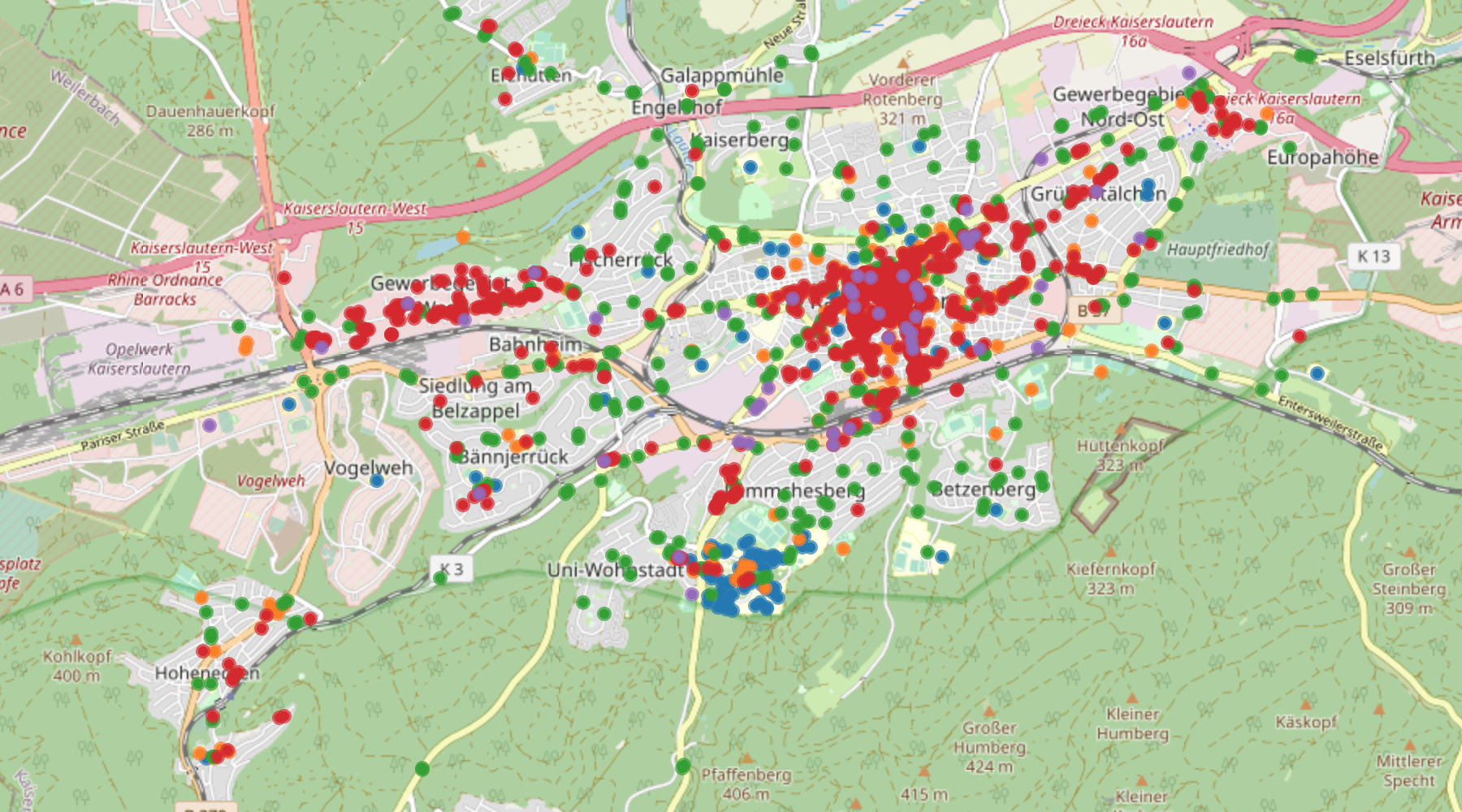}
            }
            &
            \includegraphics[width=0.38\linewidth,valign=c]{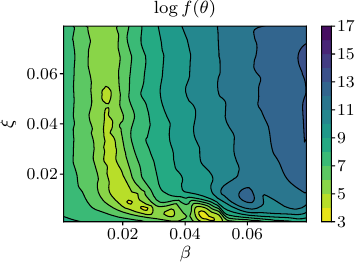}
            \\
            &
            \hspace*{-8pt}\includegraphics[width=0.33\linewidth,valign=c]{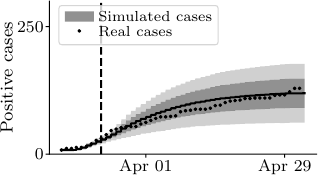}
        \end{tabular}
        \label{fig:estimation-appendix-KL}
    }
    \caption{Continued}
\end{figure}

\clearpage
\newpage

\section{Supplementary Tables}
\label{app:additional-tables}

\begin{table}[!h]
\caption{Epidemiological model parameters in units of days. Hospitalization and fatality rates $\alpha_h$ and $\alpha_b$ mentioned in the main text are estimated from COVID-19 case data in the region and is age-dependent. 
Log-normal parameters denote the underlying normal mean and standard deviation.
} \label{tab:parameters}
\centering
\begin{threeparttable}
\begin{tabular}{@{}p{\textwidth}@{}}
\centering
\begin{tabular}{lllr}
  Counting process & ~Starts when~ & ~$\log\Ncal$ parameters~ & ~Source \\\hline
  $M_i(t)$ &  $dE_i(t) = 1$ & 
  $(0.9470, 0.6669)$\tnote{$\dagger$} & \cite{lauer2020incubation} \\
  $R^{s}_i(t), R^{a}_i(t)$ & $dI^{s}_i(t) = 1$ &$(2.6365, 0.0713)$\tnote{$\ddagger$}
  &\cite{who-final-report,he2020temporal,woelfel2020clinical} \\
  $W_i(t)$ & $dI^{p}_i(t) = 1$ & $(0.7463, 0.4161)$\tnote{$\ddagger$} & \cite{he2020temporal} \\
 $Y_i(t)$ & $dI^{s}_i(t) = 1$  &$(1.9358, 0.1421)$\tnote{$\ddagger$} & \cite{wang2020clinical}\\
  $Z_i(t)$ & $dI^{s}_i(t) = 1$   & $(2.5620, 0.0768)$\tnote{$\ddagger$} & \cite{linton2020incubation} \\
  \hline
\end{tabular}
\end{tabular}
\vspace*{10pt}
\begin{tabular}{@{}p{\textwidth}@{}}
\centering
\begin{tabular}{lllr}
   &  Value & Description & Source \\\hline
  $\mu$ &  $0.55$ &relative asymptomatic transmission rate & \cite{li2020substantial} \\
  $\gamma$ & $6.3013 \, h^{-1}$ &decay of infectiousness at sites\tnote{$\mathsection$} & \cite{vandoremalen2020aerosol} \\
  $\delta$ &  $0.3654 \, h$ & non-contact contamination window\tnote{$\mathparagraph$}
  & \cite{vandoremalen2020aerosol} \\
 $\alpha_a$ & $0.4$ &proportion of asymptom. individuals & \cite{nishiura2020estimation,lavezzo2020suppression,ferretti2020quantifying} \\
  \hline
\end{tabular}
\end{tabular}
\vspace*{10pt}
\begin{tablenotes}
\footnotesize
\item[$\dagger$] Incubation period from  \cite{lauer2020incubation}, 
corrected by presymptomatic infectiousness \cite{he2020temporal}.
\vspace*{-1pt}
\item[$\ddagger$] Approximate log-normal parameters constructed because COVID-19 literature results only reported using mean or median time estimates.
\item[$\mathsection$] Assumes that transmission decays with a half-life 10 times shorter than estimated for aerosols under laboratory conditions.
\item[$\mathparagraph$] For computational purposes, set from $\gamma$ by the time when rate of transmission drops below 10\% after leaving a site.
\end{tablenotes}
\end{threeparttable}
\end{table}

\vfill 

\begin{table}[!h]
\caption{Assumed mean number of visits per week per site type by individuals of different age groups for our event-based gravity mobility model~\cite{zipf1946p}. See Section \ref{sec:exp-setup}.
}\label{tab:mobrate}
\centering
\begin{tabular}{cccccc}
Age group & Education  & Social & Transport.  &  Work &  Groceries \\
\hline
0-4 & 5 & 1 & - & - & - \\
5-14 & 5 & 2 & 3 & - & - \\
15-34 & 2 & 2 & 3 & 3 & 1 \\
35-59 & - & 2 & 1 & 5 & 1 \\
60-79 & - & 3 & 2 & - & 1 \\
80+ & - & 2 & 1 & - & 1 \\
\hline
\end{tabular}
\end{table}

\vfill 

\begin{table*}[!h]
\caption{
Summary and estimated parameters for towns and regions studied in Germany and Switzerland. 
Recall that $\beta$ denotes the individual transmission rate at public sites and $\xi$ in households estimated as described in Sections \ref{sec:sampling-inference} and \ref{sec:exp-setup}.
}\label{tab:exp-summary}
\centering
\begin{threeparttable}
\begin{tabular}{@{}p{\textwidth}@{}}
\centering
\begin{adjustbox}{max width=\linewidth}
\begin{tabular}{l l r r c c c c}
      \centering
        Region & Country & $|\Vcal|$ & $|\Scal|$ & Estimation period\tnote{$\dagger$}  & Lockdown & $\beta$ & $\xi$\\\hline
        Bern                  & CH & 133,790 & 2,174 & 03/06 - 05/10 & 03/16  & 0.0337 & 0.0038 \\
        T\"ubingen            & GER &  90,539 & 1,446 & 03/12 - 05/03 & 03/23 & 0.0402 & 0.0664 \\
        Canton Jura           & CH &  73,416 &   729 & 03/09 - 05/10 & 03/16  & 0.0131 & 0.0080 \\
        Rheingau-Taunus-Kreis & GER & 187,163 & 2,352 & 03/10 - 05/03 & 03/23 & 0.0010 & 0.0500 \\
        Kaiserslautern       & GER & 104,044 & 1,525 & 03/15 - 05/03 & 03/23  & 0.0279 & 0.0061 \\
      \hline
\end{tabular}
\end{adjustbox}
\end{tabular}
\vspace*{10pt}
\begin{tablenotes}
\footnotesize
\item[$\dagger$] Chosen such that a given region had approximately five to ten confirmed COVID-19 cases, allowing for non-degenerate and comparable initial conditions. Dates are in 2020.
\end{tablenotes}
\end{threeparttable}
\end{table*}

\end{document}